\documentclass[final]{elsarticle}

\usepackage{lineno,hyperref}
\usepackage{graphicx}
\usepackage{amssymb}
\usepackage{moreverb,url}
\usepackage{graphicx}
\usepackage{subcaption}
\usepackage{balance}
 \usepackage{amsmath}
 \usepackage{multirow}
 \usepackage{adjustbox}
 \usepackage{float}
 \usepackage[table,xcdraw]{xcolor}
\usepackage[ruled,linesnumbered]{algorithm2e}
\usepackage[british]{babel}

\modulolinenumbers[5]

\begin{document}

\begin{frontmatter}

\title{What went wrong?: Identification of Everyday Object Manipulation Anomalies}
\tnotetext[mytitlenote]{This research is funded by a grant from Istanbul Technical University Scientific Research Projects (ITU-BAP), Grant No. 40240. This research is also partially funded by Scientific and Technological Research Council of Turkey (TUBITAK), Grant No. 115E-368.}

\author[]{Dogan Altan\corref{cor1}}
\cortext[cor1]{Corresponding Author}
\ead{daltan@itu.edu.tr}
\author{Sanem Sariel}
\ead{sariel@itu.edu.tr}
\address{Artificial Intelligence and Robotics Laboratory (AIR Lab) \\ Department of Computer Engineering \\ Istanbul Technical University \\ Istanbul, Turkey}




\begin{abstract}
Extending the abilities of service robots is important for expanding what they can achieve in everyday manipulation tasks. On the other hand, it is also essential to ensure them to determine what they can not achieve in certain cases due to either anomalies or permanent failures during task execution. Robots need to identify these situations, and reveal the reasons behind these cases to overcome and recover from them. In this paper, we propose and analyze a Long Short-Term Memories-based (LSTM-based) awareness approach to reveal the reasons behind an anomaly case that occurs during a manipulation episode in an unstructured environment. The proposed method takes into account the real-time observations of the robot by fusing visual, auditory and proprioceptive sensory modalities to achieve this task. We also provide a comparative analysis of our method with Hidden Markov Models (HMMs) and Conditional Random Fields (CRFs). The symptoms of anomalies are first learned from a given training set, then they can be classified in real-time based on the learned models. The approaches are evaluated on a Baxter robot executing object manipulation scenarios. The results indicate that the LSTM-based method outperforms the other methods with a 0.94 classification rate in revealing causes of anomalies in case of an unexpected deviation. 
\end{abstract}


\end{frontmatter}


\section{Introduction}

The use of service robots in domestic environments \cite{Ersen2017} raises some important ethical concerns which must be taken into account in both robot hardware and software designs \cite{Lin2011,Grinbaum2017}. One crucial concern is execution safety in such environments where robots are in close contact with everyday objects and humans in unstructured settings. Various anomalies may result in unsafe situations which must be avoided if possible. But in some cases, anomalies are unavoidable. When anomalies are unavoidable, the system is forced to make a decision. There exists a study for self-driving cars \cite{Awad2018} where the decisions given by the car are questioned by surveying people all around the world to construct a moral machine. In this study, people are asked questions about ethical decisions when an accident is unavoidable. The results of this study show that there are no certain laws for all cultures. The situation may be different for everyday object manipulations. Certain safety criteria can be defined for interactions with objects. However, undesired situations are still likely to occur even when strict laws or regulations exist. 
When such cases occur, we at least expect from the robots to identify and explain the reasons behind these cases for ensuring safety in further scenarios or realizing their limits in unforeseen settings. 
This is the main motivation behind our work. We investigate such onboard anomaly identification and inference methods to explain circumstances on unsafe tabletop object manipulation scenarios. This is mainly important to prevent further potential damages in the environment.

\begin{figure*}[t]
\captionsetup[subfigure]{}
    \centering
    \begin{subfigure}[t]{0.3\textwidth}
        \centering
        \includegraphics[width=1.34in]{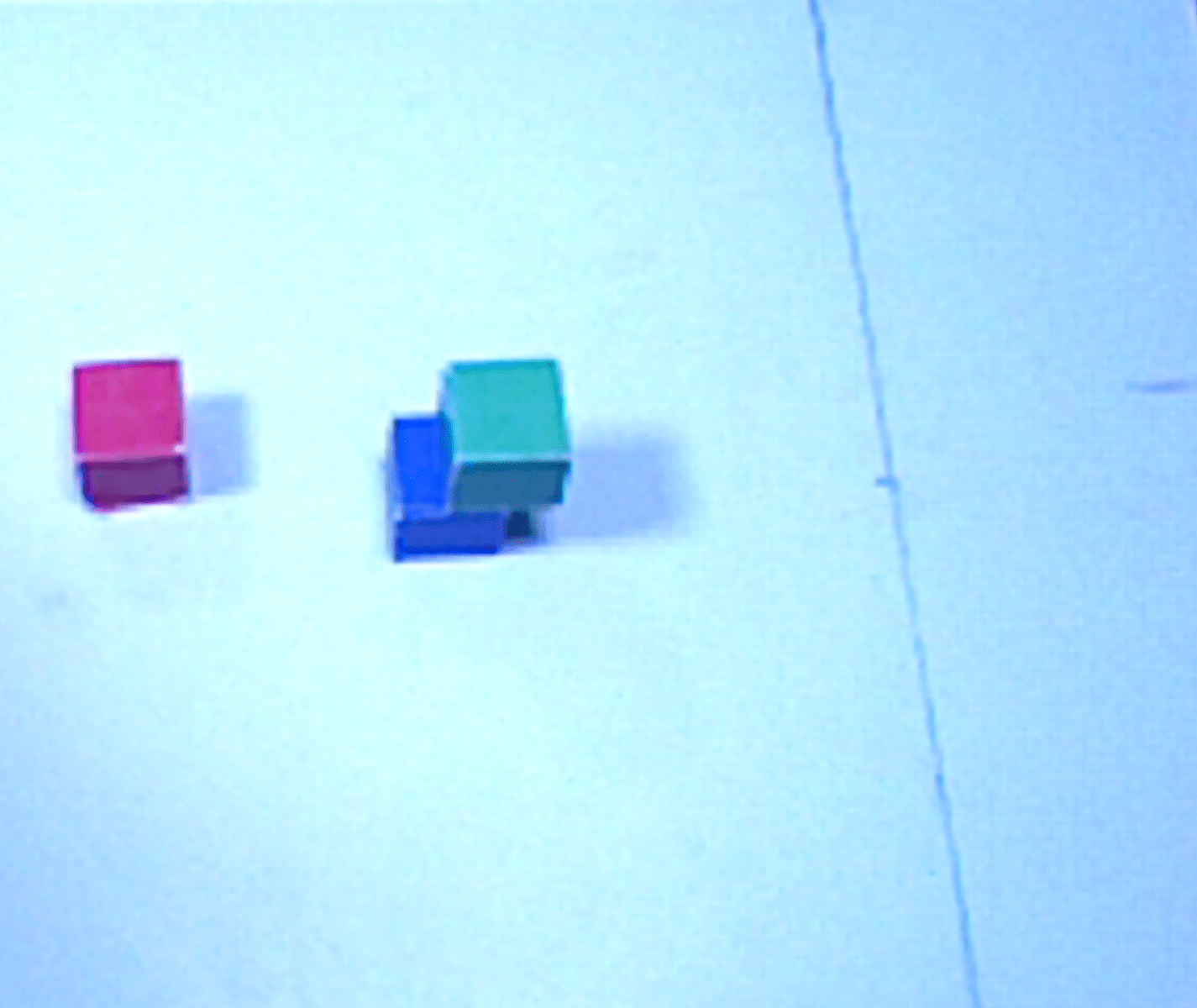}
    \end{subfigure}%
    ~ 
    \begin{subfigure}[t]{0.3\textwidth}
        \centering
        \includegraphics[width=1.34in]{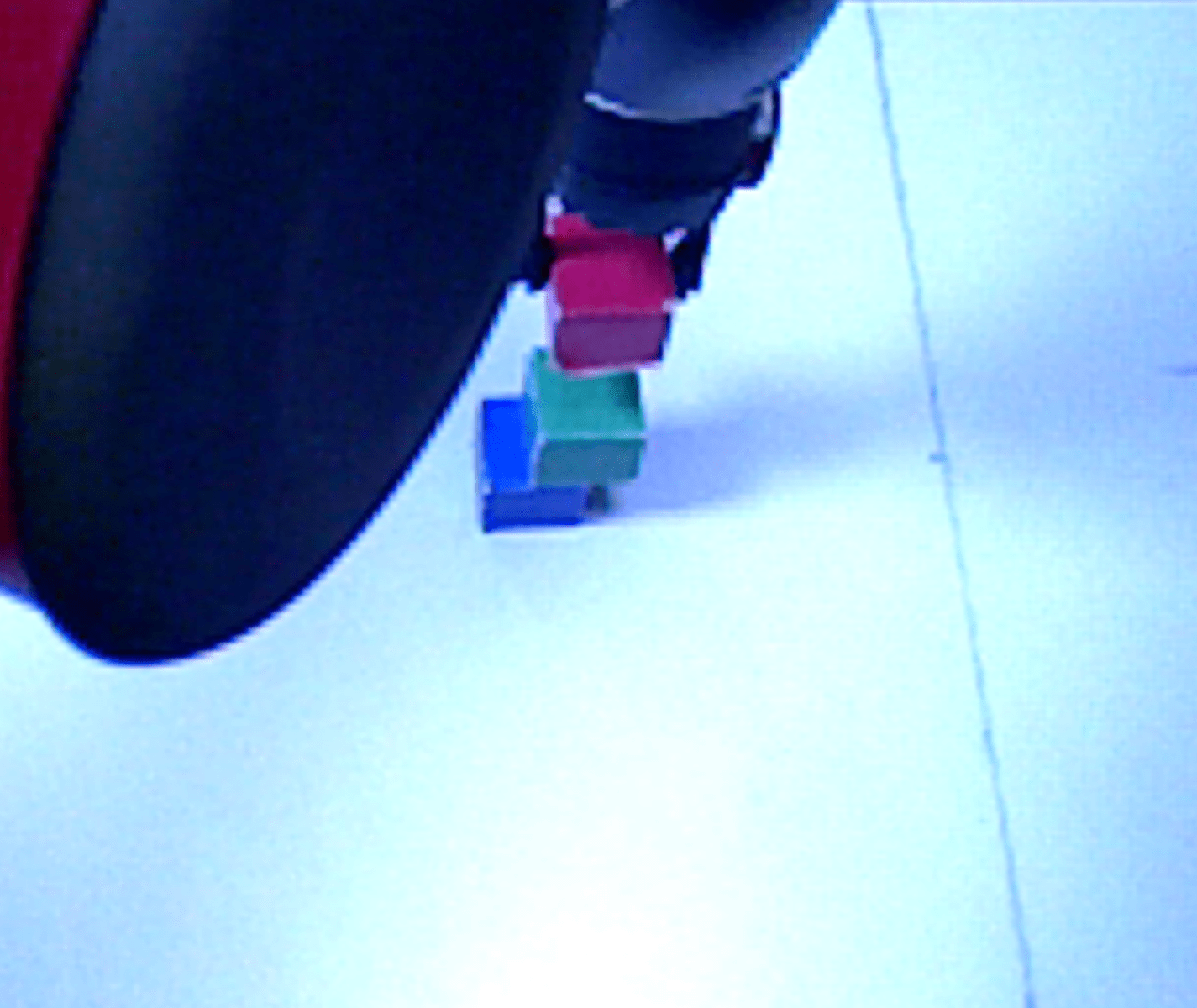}
    \end{subfigure}%
     ~
    \begin{subfigure}[t]{0.3\textwidth}
        \centering
        \includegraphics[width=1.34in]{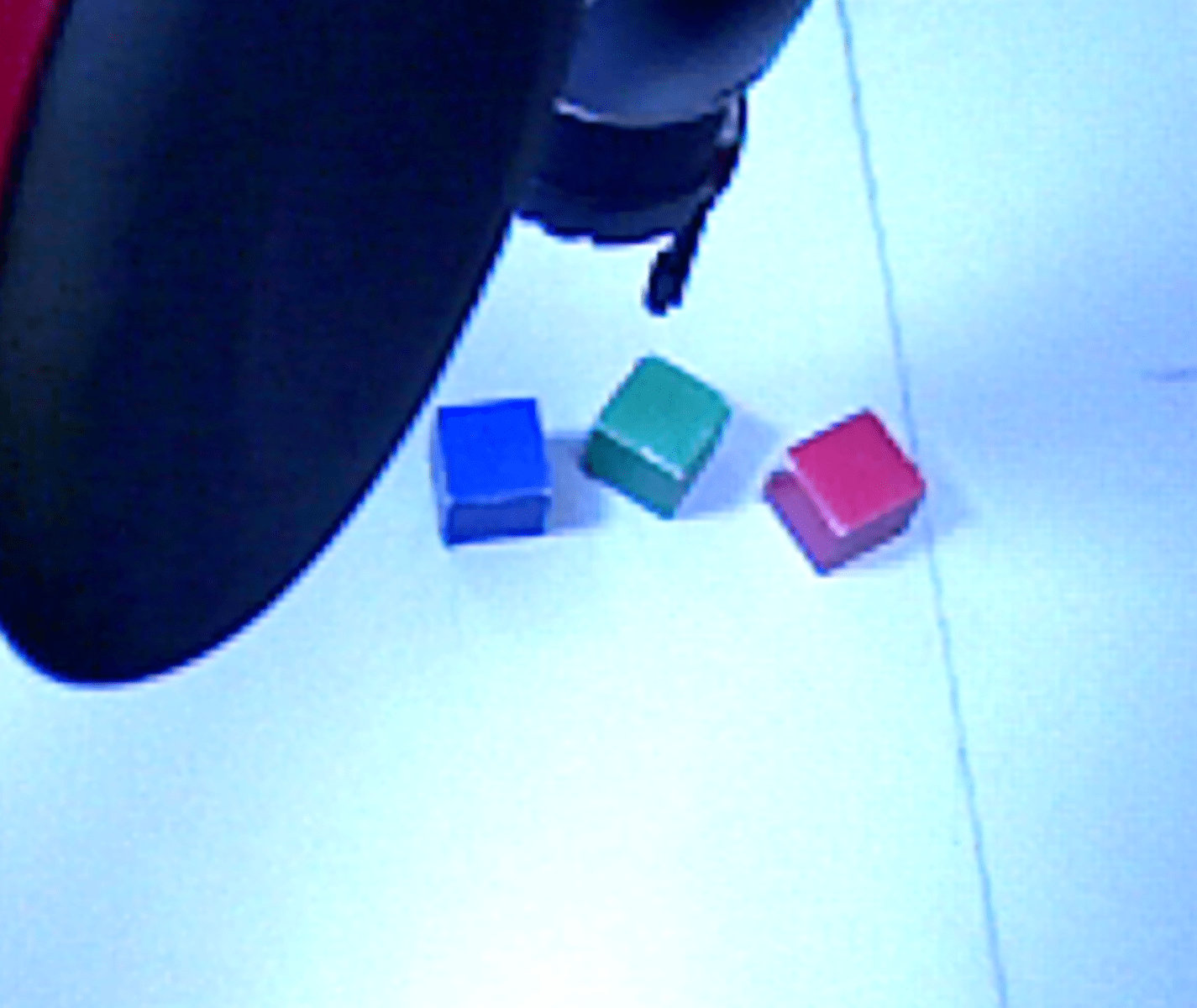}
    \end{subfigure}
    \\
     \begin{subfigure}[t]{1\textwidth}
        \centering
        \includegraphics[scale=0.62]{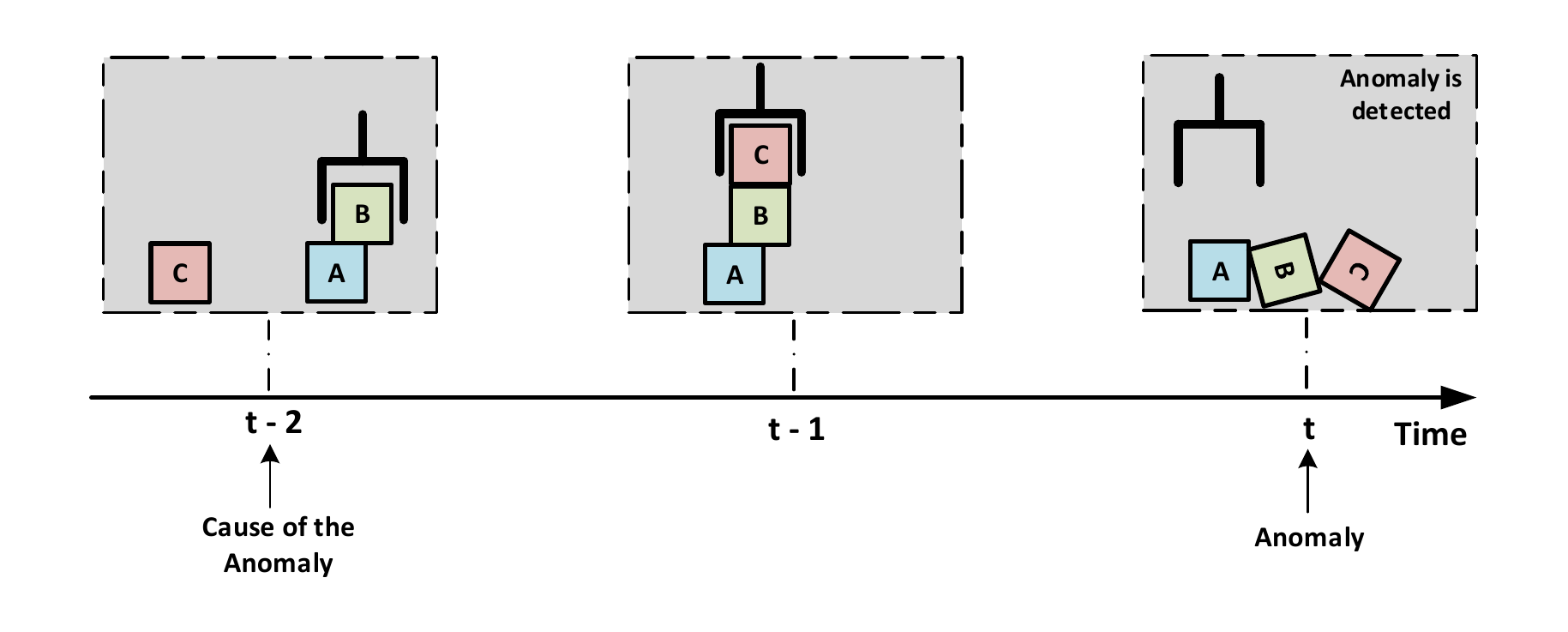}
    \end{subfigure}
    \caption{\label{tower-anomaly}  An anomaly case where a 3-block tower  collapses after the robot places the last block.  At time step $t - 2$, the robot first places block B on block A. Then, at time step $t - 1$ block C is placed on top of block B. At time step $t$, the tower collapses. The anomaly is detected at this time step. However, the real cause of this anomaly is due to an action executed in an earlier time step ($t - 2$).}
\end{figure*}





Safe execution is one of the most important  primary concerns for robot ethics. These concerns must be fully taken into account in both  robot software and hardware designs. It is better to design robots that they completely prevent anomalies. However, complete prevention may not be possible for some cases. To still recover from such a case, robots must be equipped with skills to recognize unsafe situations and their reasons. 
Recognizing and recovering anomalies resulting from either unexpected outcomes of actions or external factors requires a diagnostic procedure 
including three main steps: detection, identification and recovery \cite{Isermann1997}. First, the cases when the desired effects of actions are not met must be determined. This corresponds to the anomaly detection phase, and a continuous monitoring procedure is needed for onboard detection. After then, an identification procedure is needed for revealing and explaining the reasons of the detected anomalies. Then, as a last phase, the system must be adapted to a nominal state as a recovery. In this study, we mainly focus on the identification problem for the reasons of anomaly cases after they are detected.

Identification and inference procedures must relate anomalies to potential causes. Some example causes include an error in the software of the robot (e.g., a misclassification of objects by the vision system), a malfunction of an hardware part (e.g., gripper), improper task execution (e.g., unstable placements) or other external factors (e.g., displacement of target objects by other agents) \cite{Altan2014}. 
These relations must be made by a robot with an identification procedure to reveal the real cause of an anomaly by applying a time series analysis of its observations. An example scenario is given in Figure \ref{tower-anomaly} presenting an anomaly case for a block stacking scenario. In this figure, an example sequence, from the head mounted camera of a robot on the top sub-figures, is presented along with the visual illustration of the actual execution phases on the bottom. In this scenario, the robot is tasked to construct a 3-block tower from the given blocks $A$, $B$ and $C$ on the table. At time $t-2$, the robot places block $B$ on top of block $A$ with an offset, that is their center of messes are not on the same vertical axis. Then at time step $t-1$, it picks block $C$ to place it onto block $B$. After placement at time step $t$, the tower collapses since the physical balance of the tower is not satisfied. Since at time step $t-1 $, block B is not well aligned with block A, this causes an unstable placement. Although, this instability does not result in a damage at time step $t-1$, it leads further damages in later time steps. Therefore, the robot needs to determine what went wrong in which time step.



In this study, we present and analyze Long Short-Term Memories (LSTMs), Conditional Random Fields (CRFs) and Hidden Markov Models (HMMs) as anomaly identification procedures. Outputs of auditory, visual and proprioceptive sensor modalities are fused, and the fused historical observation data are investigated to reason about an anomaly. Based on our analysis on a number of anomaly scenarios with a Baxter humanoid robot, we show that our LSTM-based method gives better results. 

The main contribution of this study is the presentation of the anomaly identification procedures that fuse auditory, visual and proprioceptive sensor modalities in everyday object manipulation scenarios. To the best of our knowledge, this is the first time that anomaly identification procedures are proposed for unstructured environments where safety is more challenging compared to industrial settings in which task specifications are fixed and stable.

The rest of this paper is organized as follows: First, the literature on anomaly identification problem is summarized. Next, the presented problem is formulated, followed by the proposed anomaly identification methods in detail. Afterwards, a number of anomaly cases are investigated, and the methods are analyzed on these case scenarios with a Baxter robot.  Finally, the paper is concluded with discussions and possible future directions.

\section{Literature Review}

Detecting and revealing the underlying reasons behind anomaly cases is investigated commonly by researchers in the literature \cite{Pettersson2005,Fritz2005}. This problem is particularly important due to the needs for safety in robotics. Possible anomalies for robotic environments are classified in taxonomies \cite{Carlson2004} and \cite{Karapinar2012}. Anomalies can be classified under two categories by considering their sources: physical and human \cite{Carlson2004}. The other taxonomy \cite{Karapinar2012} investigates anomalies under two main categories: internal and external. In this classification, internal anomalies correspond the anomalies related to the belief of the robot where external anomalies correspond to the environmental issues. Anomalies that may occur in smart manufacturing environments are also categorized in \cite{Lopez2017}. This section summarizes and compares the most related work to this research. 

Generally in the literature, logic-based, sensor fusion-based, expert systems-based, and probabilistic methods are investigated to reveal the causes of anomalies. In a logic programming-based study, hypotheses are constructed and maintained to explain anomaly cases \cite{Gspandl2012}. Hypotheses are attached with related costs, and they are maintained in a hypothesis pool. In case of an anomaly, inconsistencies among the hypotheses are used to identify anomalies. In another work \cite{Steinbauer2009}, an algorithm that considers inconsistencies between the theory and the model of the world is presented to identify anomaly cases. In a multi-robot domain, a cooperative anomaly diagnosis method is presented where robots are able to help each other in diagnosing the anomaly case by using an agent programming language \cite{Morais2015}.

\cite{Abid2015} proposes a multi-level sensor fusion to detect and identify abnormal cases by clustering sensors and processing their outputs. This method is tested on navigation scenarios with a mobile robot. Another clustering based method uses Global Fuzzy C-means Clustering Algorithm to construct clusters to find the causes of the failures \cite{Mendoza2015}. After detecting the failure with clustering, identification is achieved by considering the gathered sensor values and the closest clusters. 

A data-driven method is also presented to switch and select the most appropriate successive behavior by processing sensor modalities such as haptic, auditory and visual \cite{Kappler2015}. \cite{Baghernezhad2016} proposes locally linear models (LLMs) that utilize adaptive threshold bands and model error modeling (MEM) techniques to achieve anomaly detection and identification problem for a mobile robot setting.

Expert systems based on human knowledge are also presented to identify anomaly cases \cite{Nan2008}. The main drawback of this method is the high dependency of the performance to the expert knowledge. There may also be cases where a planner can not come up with a plan to achieve a goal. In \cite{Gobelbecker2010}, causal graphs and domain transition graphs are used to analyze such cases to identify why the planner fails to generate a plan. 

Dynamic Bayesian Networks (DBNs) \cite{Flores-Quintanilla2005}, Particle Filters (PFs) \cite{Verma2004} and Kalman Filters \cite{Rigatos2009} are also investigated as probabilistic structures in order to handle anomaly cases. A variety of Hidden Markov Models (HMMs) are investigated to detect and identify anomalies in assistance tasks \cite{Park2018HMM}. In an HMM-based approach, the problem of skill and anomaly identification is studied with gradient analysis \cite{Luo2018}. In this study, scenarios that include pick and place actions are analyzed with a humanoid robot. Bayesian Filters are also proposed in an industrial robotic domain to analyze unexpected deviations \cite{Di2013}.  

Convolutional Neural Networks (CNNs) are used to construct a model for task execution in a simulation environment \cite{Bowkett2018} by using visual features on synthetic depth images. Another deep learning based anomaly identification method presents a solution to the problem in the domain of human-robot interaction \cite{Park2018LSTM}.

Our previous works \cite{Altan2014,Altan2016Empirical} present probabilistic approaches to reveal the causes behind anomalies for mobile robots. Hierarchical Hidden Markov Models (HHMMs) and Particle Filters (PFs) are processed in parallel to achieve multi-hypo-thesis tracking in mobile robot manipulation scenarios. In this work, our main objective is identifying anomalies in long-term tabletop manipulation scenarios. Therefore, there is a need of a more efficient temporal analysis which can be tackled by LSTMs.

\section{Anomaly Identification Problem}
In this section, we formulate the anomaly identification and inference problem and present case scenarios we analyze and potential anomalies that may occur during them. 
\subsection{Problem Formulation}

Formally, assume that the robot has a plan $\pi$ which is an of a $k$ action sequence $\pi = \{a_0, a_1, ... , a_k\}$ (given by an operator or derived by a planner), and let $\mathcal{A}$ be the set of all actions that the robot has. 
During action execution, it needs to observe the real world through its sensors. Assume the robot has $m$ sensors to sense the environment, and the sample gathered from a particular sensor modality at time $t$ is denoted with $s_m^t$. An observation at a particular time step can be represented as $x_t$ where $t$ indicates the time step that the corresponding sample is gathered. Each observation is a combination of the data gathered from the sensors of the robot, and can be represented as a tuple, $x_t = (s_0^t, s_1^t, ..., s_m^t)$. Therefore, an execution sequence can be represented as a sequence of observations. 

While the robot executes its actions, it senses the environment continually and constructs a sequence of the observations that are gathered. Denoting an execution sequence (observation history) with $S$, a sequence of observations can be written as follows:
\begin{center}
    $S = \{(x_0,y_0), (x_1,y_1), ...... , (x_t,y_t)\}$
\end{center}

Each element of this sequence is a pair: observation ($x_t$) and hidden state of the observation ($y_t$). $y_t$ corresponds to the ground truth for the state, and this state is hidden to the robot. The domain of $y$ is composed of anomaly cases in which unexpected outcomes occur, $y \in F$. The overall anomaly set can be given as $F = \{anomaly_0, anomaly_1, ... , anomaly_n\}$ where each variable in this set corresponds to a distinct anomaly case and $n$ is the number of anomaly classes.

After the robot detects an anomaly case, it should be able to decode the history and label the states corresponding to the observation sequences in order to explain the reason of the anomaly case. This can be achieved by assigning probabilities to anomaly cases for each execution state at time $t$. 
In the end, it should come up with the related labels for each execution state $y_t$ in an execution sequence $S$. This task can be formalized as follows: 

\begin{center}
\begin{equation}
argmax_i(P(y_t = anomaly_i | x_{0:t}, \mathcal{A}))
\end{equation}
\end{center}

\subsection{Investigated Manipulation Actions}
We investigate the anomaly identification problem under the assumption that the robot is able to execute the following primitive actions to interact with the environment and the objects:  \textit{move-towards-object, move-to-loc, pick-up}, \textit{put-down}, \textit{put-down-on} and \textit{push}. These actions described as follows:

\begin{itemize}
\item \textit{move-towards-object(objectX)}: The robot moves its arm to an the location of objectX.
\item \textit{move-to-location(destination)}: The robot moves its arm to a destination.
\item \textit{pick-up(objectX)}: The robot picks objectX with its gripper.
\item \textit{put-down(objectX)}: The robot releases objectX, that is currently being hold, from its gripper to the table.
\item \textit{put-down-on(objectX, objectY)}: The robot releases objectX, that is currently being hold, from its gripper on top of another objectY.
\item \textit{push(objectX, axis, distance)}: The robot pushes objectX along a given axis for a specified distance.

\end{itemize}

\subsection{Scenarios and Potential Anomaly Cases}
We focus on a set of anomalies that may occur on the selected action set. The following scenarios list some case scenarios that we analyze:

\begin{itemize}
\item The robot is tasked to push an object along an axis for a given amount. 

\item The robot is tasked to pick an object up with its gripper.

\item The robot is tasked to build a tower from a set of cubic blocks.  

\end{itemize}

While the robot runs these scenarios, miscellaneous anomaly cases may occur. These anomalies are listed as follows:
\begin{itemize}
    \item \textit{A target object is not in its previous location but in a different location (LOC)}: This anomaly case stands for the cases where a target object is displaced by an external agent or human.
    \item \textit{A target object can not be pushed or grasped due to the disappearance of it (DIS)}: This anomaly case corresponds to the situations where a target object to be manipulated and which was initially in the field of view of the robot, is taken out of the field by an external agent. These kind of situations lead the robot to an anomaly state since the action requires an interaction with that particular object. 
    \item \textit{The built structure is collapsed (UNB)}: This anomaly case corresponds to the situations where a structure, particularly a tower of objects collapse due to an instability or unbalance during construction.
\end{itemize}

\begin{table*}[!ht]
\begin{adjustbox}{max width=\textwidth}
\begin{tabular}{|>{\centering\arraybackslash}m{2.4cm} | >{\arraybackslash}m{5cm}  |c| >{\arraybackslash}m{5.8cm}  |}
\hline
\textit{\textbf{Anomaly}}               & \multicolumn{1}{c|}{\textit{\textbf{Explanation}}}                                                                              & \textit{\textbf{Action}} & \multicolumn{1}{c|}{\textit{\textbf{Example Scenarios}}}                                                                                                                                                             \\ \hline \hline

A target object is not in its previous location but in a different location (LOC)       & An object's location is changed by an agent without the knowledge of the robot.                                                 & all actions                           & - While the robot approaches an object to manipulate (grasp, push etc.) it, the location of the object is changed. - While the robot constructs a tower from a set of objects, object locations are changed.                                                                                    \\ \hline 
A target object disappears which was previously in the field of view (DIS) & An object, that was in the environment, is taken out of the environment by an agent.                                            & all actions                           & - While the robot approaches an object to grasp or push it, the object is taken out of the environment. - While the robot constructs a tower from a set of objects, one or more objects are taken out of the environment. \\ \hline
The built structure is collapsed (UNB)         & While the robot constructs a structure from a set of objects, due to an misaligned sub-structure in a previous time step, the current structure collapses. &     put-down-on                       & The robot tries to construct a tower from some objects. Due to an unstably placed object in a previous action, the structure collapses.                                                                        \\ \hline
\end{tabular}
\end{adjustbox}
\caption{\label{anomaly-dict} Dictionary of the anomalies that are handled.}
\end{table*}

Table \ref{anomaly-dict} presents an overview of the anomalies that are explained in this section. Each row of the table corresponds to an anomaly type. The first column represents the name of the corresponding anomaly followed by the explanation of the anomaly. The next column explains the actions during which the corresponding anomaly may occur. Finally, the last column presents example scenarios for that kind of anomaly that are investigated in this study. Note that, as seen from the table, each anomaly type may occur in any of the given example scenarios. For example, while constructing a tower, an object in the environment may disappear, its location may be changed or the tower may collapse.

\section{Anomaly Identification}
In this section, we first explain the underlying hardware and software structure for detecting anomalies which directly contributes to the anomaly identification problem. It is important to present the low-level procedures and our hardware setup to clarify the anomaly identification procedure on top of them. We then present our Long-Short Term Memories (LSTM) based anomaly identification procedure. Furthermore, the implementation of Hidden Markov Models (HMM) and Conditional Random Fields (CRF) for anomaly identification are also presented. These methods are selected for comparative assessment of our method. 

\subsection{Hardware Setup and Low-level Processes to Detect Anomalies}
\begin{figure}[h]
    \centering
        \includegraphics[scale=0.055]{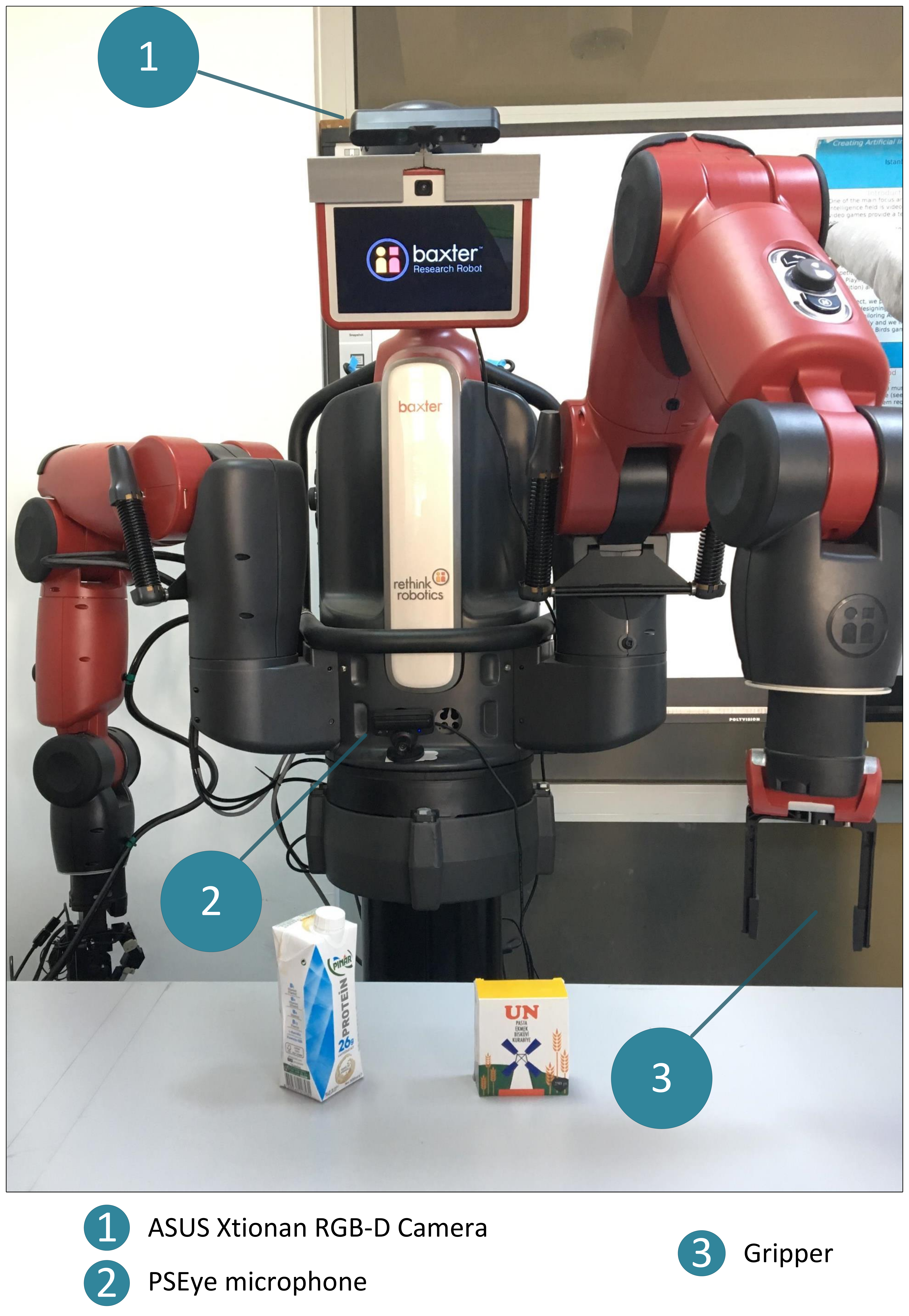}
    \caption{\label{baxter} The Baxter humanoid robot which can manipulate objects and detect anomalies in a tabletop environment.}
\end{figure}

Our case setup includes a Baxter humanoid robot (Figure \ref{baxter}) that can run the explained scenarios. 
The robot is equipped with a gripper attached to its arm to manipulate objects, a head mounted Asus Xtion RGB-D camera to collect visual clues and a PSEye microphone attached to its body to process the sound data. The robot framework is constructed and integrated on Robot Operating System (ROS) \cite{Quigley2009} that provides libraries and tools for robotic applications.

In order to identify anomalies, first, they need to be detected by the robot. The robot should possess a procedure that continually monitors the execution and evaluates the observations to decide on whether there is an anomaly or not. To do so, the robot needs to fuse various sensor modalities and interpret them to decide on an anomaly case. 
We use an existing anomaly detection procedure \cite{Inceoglu2018b,Inceoglu2018Failure}
based on an HMM-based algorithm that runs on the fused data.  
The following subsections present these sensor modalities and how their data are processed for both detection and identification.

\subsubsection{Audition}
\label{sound}
A microphone is placed on the body of robot to keep track of sound data during execution. It gathers sound data through four channels and the data on one of these channels are processed during the execution. Discrete Fourier Transform (DFT) is used to convert sound data to the frequency domain. Then, Mel Frequency Cepstral Coefficients (MFCC) algorithm \cite{Han2006} is invoked to detect sound information and its duration. With this algorithm, fragments that include sound data with exceeding an emprically-defined threshold value are classified by a Support Vector Machine (SVM)  \cite{Inceoglu2018Failure,Saltali2016}.

\subsubsection{Vision}
LINEMOD \cite{Hinterstoisser2012} and LINEMOD-HS algorithms \cite{Ersen2013} are used for object recognition. Objects are recognized with the data gathered through the RGB-D camera which provides depth information of the objects with color information. The templates of the objects are stored in the knowledge base (KB) of the robot and during the execution, point clouds from the scene are compared with the templates in the knowledge base of the robot. If a point cloud is matched with a template, the object is recognized with a similarity measure \cite{Inceoglu2018}. 

A depth-based segmentation algorithm is also run for detecting point clusters (segments) of the scene beside the template-based object recognition algorithm. Organized Point Cloud Segmentation \cite{Trevor2013} is used to extract clusters/segments from the scene gathered with the RGB-D camera. Clusters that are larger or smaller than a predefined threshold are discarded. Note that the segments clustered with the segmentation algorithm are classified as unknown objects, these segments do not have an object type \cite{Inceoglu2018}.

All these vision sources are fused together to construct the world model consistently by the Violet interpreter system \cite{Inceoglu2018}. Violet continually monitors the scene and fuses the data in order to update the world model of the robot. For example, when an object is added to the scene, it updates the representation of the robot's world accordingly. It is also capable of extracting spatial symbolic relations among the objects in the environment such as \textit{on} and \textit{in}.

\subsubsection{Proprioception}

The gripper state of the robot is used as a proprioceptive information in this study. The distance between the gripper fingers and the force measurements gathered from the gripper are processed. 

\subsection{Features for Anomaly Identification}

During the training phase of the anomaly models, various features are used to represent an observation of the robot. These features are itemized and explained as follows:

\begin{itemize}
\item  \textit{Laser Distance: } This feature specifies the distance of the gripper from the object to be manipulated. The laser is mounted on the gripper of the robot.

\item \textit{Gripper State: } This feature specifies the state of the gripper of the robot. In can be either \textit{open} or \textit{close}. Note that the range of the gripper is 8 centimeters. Depending on the size the object at hand, the robot adjusts its gripper to hold an object. 

\item \textit{Gripper Force: } This feature specifies the force value at the gripper. It can measure force on a scale of 0 to 35 Newton. After pressure is detected, the sensor produces a normalized value that differs between 0 and 100 depending on the corresponding torque value.

\item \textit{Sound: } This feature specifies the sound information after it is classified. The sound information is gathered through the microphone of the robot, and there are 3 classes: \textit{no sound, drop} and \textit{ego noise} \cite{Inceoglu2018Failure}. \textit{No sound} represents the situation where no sound is received. \textit{Drop} class corresponds to the sound of a fall of an object in the environment, where \textit{ego noise} represents the sound of the robot itself.

\item \textit{Object Information: } This feature corresponds to the belief of the robot to the existence of the objects in the environment. It includes the information related to the objects including their locations and relative positions among each other (e.g. the offset of the objects when there is an on relation between them). There are three classes for the existence of an object: yes, no and unknown. \textit{yes} indicates that the object is recognized by the vision module, and it is observed by the robot where the value \textit{no} indicates the non-existence of the object. \textit{unknown} denotes that the visual updates for that object is suspended due to an occlusion with the robot's arm.

\item \textit{Action Phase: } This feature corresponds to the phase of the action that the robot executes. For instance, for action \textit{pick-up}, there are 5 phases such as: \textit{pick-started, pick-approaching-object, pick-gripper-open, etc.}. The actions \textit{push}, \textit{put-down} have 6 phases where \textit{move-to-object} and \textit{move-to-location} include 5 phases.

\end{itemize}

\subsection{LSTMs for Anomaly Identification}

Long Short-Term Memory (LSTM) \cite{Hochreiter1997} is a special structure of Recurrent Neural Network (RNN) that overcomes the problem of modeling long dependencies. RNNs are able to propagate historical information from the past, however, whenever the interval size increases, the information that needs to be transferred is vanished, and this problem is called vanishing gradients problem. Therefore, LSTMs alleviate this problem by representing a complex and derived version of the repeating module that exists in RNNs as well. However, instead of having a one-layered neural network layer, it has four layers. 

Generally, the idea behind an LSTM is straightforward. An LSTM cell is a unit that flows the information from the entrance through the exit by applying necessary modifications on it. The LSTM structure modifies this cell to remove or add information. Simply, there are four phases for modifying this cell: removing unnecessary information, adding new information, updating the previous memory cell and producing the output.

 The equations that summarize these operations for an LSTM are given below \cite{Hochreiter1997}. 
 
\begin{align}
f_t = \sigma(W_f y_{t-1} + U_f x_t + b_f) \label{eq:forget}\\
i_t = \sigma(W_i y_{t-1} + U_i x_t + b_i) \label{eq:select1}\\
\widetilde{c_t} = tanh(W_c y_{t-1} + U_c x_t + b_c) \label{eq:select2}\\
c_t = f_t \odot c_{t-1} + i_t \odot \widetilde{c_t} \label{eq:update}\\
o_t = \sigma(W_o y_{t-1} + U_o x_t + b_o) \label{eq:output1}\\
y_t = o_t \odot tanh(c_t) \label{eq:output2}
\end{align}

Note that in the equations, $y_t$ corresponds to the hidden variables at time $t$, $x$ is for input variables, $b$ is for bias for the corresponding variables, $W$ and $U$ denote weights for the hidden states and inputs, respectively. $c$ denotes cell state, where $f$ is forget ratio and $i$ is selected input variables. $\odot$ is element-wise multiplication. In order to decide what is stored and forgotten in the cell, a sigmoid function is implemented. This function simply gives the degree of the forget operation. In case of a higher value that the function returns, more information is stored in the cell from the previous state (Equation \ref{eq:forget}). After deciding the degree of removal for the cell, the information that is going to be stored in it should be decided. This is achieved by selecting the variables (Equation \ref{eq:select1}) and constructing new candidates for the update (Equation \ref{eq:select2}) with sigmoid and tangent functions, respectively. Later on, the cell state is updated (Equation \ref{eq:update}) and the output is produced (Equations \ref{eq:output1} - \ref{eq:output2}). 

\begin{figure*}[h!]
        \centering
        \includegraphics[width=4.9in]{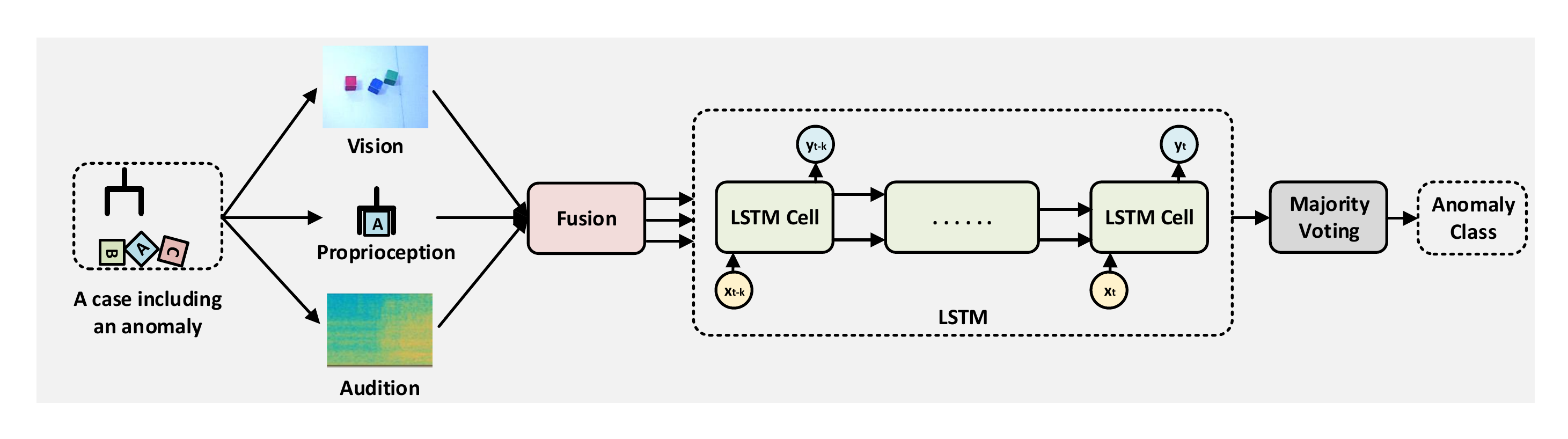}
    \caption{\label{lstm-framework} An illustration of the proposed anomaly identification method. It fuses various sensor modalities to identify anomalies.}
\end{figure*}

Figure \ref{lstm-framework} illustrates the overall LSTM-based identification procedure. First, an anomaly model is constructed by a training algorithm with a data set of anomalies that includes anomaly cases. After constructing a model with a suitable parameter learning algorithm, it can be used to identify an anomaly. The proposed method accepts an anomaly model, and a plan to achieve the given task as a sequence of actions. Later on, while there is an action in the plan to be executed, it is executed continually unless an anomaly case is detected. Observations that are gathered through the sensors are fused and maintained as observation history to be examined later in case of an anomaly case. Whenever an anomaly is detected, the execution sequence that is maintained is investigated by the trained model. After assigning an anomaly label to the execution states, majority voting is applied to the whole  sequence to find out the cause of the anomaly.

\begin{algorithm}[h]
 \KwData{Trained anomaly model $M$, Plan to be executed $\pi$}
 \KwResult{Anomaly class $y$}
 $S$ = $\emptyset$\;
 \While{$\pi \neq \emptyset$ and !anomalyOccurred}{
  $a$ = POP($\pi$)\;
  \While{!anomalyOccurred and !actionOver(a)}{
  $s_v{^t}, s_a{^t}, s_p{^t}$ = \textit{sense-the-environment()}\;
  $s_v{^t}$ = \textit{sceneInterpretation()}\;
  $s_a{^t}$ = \textit{classifySound()}\;
  $s_p{^t}$ = \textit{tactileForce()}\;
  $x_t$ = \textit{sensorFusion($s_v{^t}$,$s_a{^t}$,$s_p{^t}$)}\; \label{fusion}
  $S$.\textit{append($x_t$)}\; 
  $anomalyOccurred$ = \textit{anomalyDetection($x_t$)}\; 
  }
 }
 \eIf{$anomalyOccurred$}{
    \For{each time step $t$}{
    $y_t$ = \textit{identifyAnomaly($M$, $x_t$)}\tcp*{$S = (x_t, y_t)$}}   }{
   $y_{0:t}$ = \textit{safe}\;
  }
  $y$ = \textit{majorityVoting($y_{0:t}$)}\;
  \Return $y$
 \caption{Anomaly Identification Algorithm}
 \label{algo}
\end{algorithm}

Algorithm \ref{algo} presents an overview of the proposed anomaly identification algorithm in detail. The algorithm accepts a model that is previously trained ($M$) and a plan to be executed ($\pi$). As an output, it provides the class of the anomaly ($y$) explaining the unexpected situation. Initially, the observation sequence ($S$) is empty. First, the robot starts to execute the actions in the plan one by one. During this phase, if everything goes as planned, the robot observes the environment by gathering data through its sensors. $s_v{^t}$ denotes an observation gathered from the visual sensor where the subscript is the sensor modality and $t$ is the time of the observation. Similarly, $s_a{^t}$ denotes the observation gathered from the auditory sensor, and $s_p{^t}$ are the data gathered from the gripper. After gathering these data, they are processed and fused. First, visual data are interpreted with Violet \cite{Inceoglu2018}, and the world model is constructed. Then, the auditory data are classified if a sound is received with the method that is explained in Section \ref{sound}. Last, tactile force and laser data are collected. These data need to be fused since they may not be synchronous. Line \ref{fusion} fuses these data in order to construct a single observation tuple $x_t$ for time step $t$. Therefore, every time step $t$ includes an observation of each corresponding sensor modality after the fusion procedure. Later on, the observation $x_t$ is maintained in observation history $S$. 

Unless an anomaly is detected, this procedure continues. In case of an anomaly, it is identified by using the observation history $S$ and the anomaly model $M$. The observation sequence is fed into the trained model, and each observation is labelled with one of the anomaly classes or the label \textit{safe} with the trained model. Note that at the end of this procedure, a label sequence is constructed. Therefore, an overall decision on a label is required. In order to end up with a consensus among the labelled observations for the decision, majority voting is applied on the labelled sequence. The anomaly class with the most votes in the sequence is selected as the identification of the anomaly.

\subsection{HMMs for Anomaly Identification}

\begin{figure}[h]
\begin{center}
     {\includegraphics[scale=0.52]{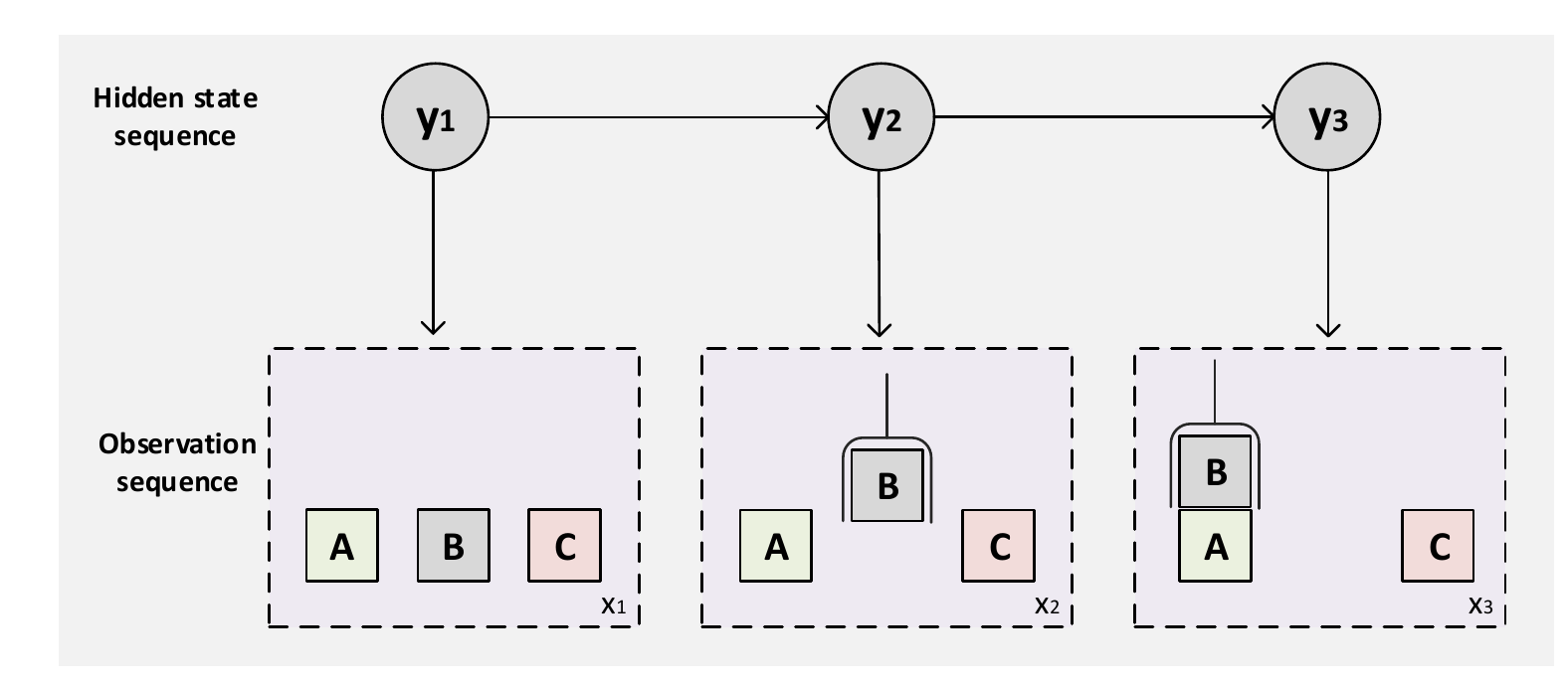}}
\caption{\label{fig:hmm-scenario}An example sequence for an HMM for the object stacking scenario.}
\end{center}
\end{figure}

HMMs \cite{Baum1966} are generative temporal probabilistic models and their underlying structure is Markov processes. Simply, an HMM consists of five components: hidden states ($y_i$), observations ($x$), transition model ($A = a_{y_i y_j}$ where $y_i$ and $y_j$ are hidden states), observation model ($B = b_{y_i}(x)$ where $y_i$ is a hidden state and $x$ is an observation) and an initial state distribution ($\Pi_i$). States that can not be directly observed are called as hidden states in the structure. The model that defines the transition likelihood between the states are denoted as transition model where the probability of observing an observation in a distinct state is called observation model. The belief on the initial state of the model is defined with the initial state distribution. Figure \ref{fig:hmm-scenario} illustrates a general HMM model in a block stacking domain. In the figure, upper sequence corresponds to the hidden states where the lower part illustrates the observations that the model emit at each time step. 

In order to model the presented problem with HMMs, each anomaly type is represented as a distinct HMM, corresponding to an anomaly type. Therefore, each anomaly type is trained for the related anomaly type and is capable of applying binary classification. After an anomaly is detected, each HMM is fed with the observation sequence that is gathered through the sensors of the robot. First, the sequence is classified considering the score gathered from each HMM model. After deciding on the model that the sequence is most likely to belong, the whole sequence is labelled by the related model with the Viterbi algorithm \cite{Forney1973}. The Viterbi algorithm utilizes a Viterbi value $v_t$ and $\delta_t$ for an observation to find the most likely explanation of a given sequence. Note that in the equations, $k$ is the state index that maximizes the given statement, and $v_t$ and $\delta_t$ are calculated for each time step with the following equations.

\begin{center}

$v_t(y_i) = max_k(v_{t-1}(k) * a_{y_k y_i}) * b_{y_i}(x_t)$

$\delta_t(y_i) = argmax_k(v_{t-1}(k) * a_{y_k,y_i})$

\end{center}

In order to model the problem with HMMs, each HMM has binary hidden states: \textit{safe} and \textit{anomaly}. \textit{safe} corresponds to the case where the state does not include an anomaly, in other words, the state is safe. On the other side, \textit{anomaly} stands for the cases where the state includes an anomaly. Observations correspond to the sensory data that are gathered during the execution for each time step. 

The HMM-based procedure accepts trained HMMs and the plan to be executed by the robot. Initially, the execution history that corresponds to the observation sequence is empty. First, the plan, that is a sequence of actions, is started to be executed. While the plan is not over and an anomaly is not detected, the observations are gathered through the sensors and stacked to the history. In case of an anomaly, likelihoods are calculated to reveal the model that fits most to the observation sequence (in other words, the model that is more likely to generate the observation sequence), and the model with the highest likelihood is selected. This is followed by executing the Viterbi algorithm with the selected model on the sequence. Note that since the scope of the manuscript is anomaly identification, the labels are only assigned in case of an anomaly. Therefore, if an anomaly does not occur, the algorithm labels all the sequence with \textit{safe} label. The procedure terminates with returning the labelled sequence. 

\subsection{CRFs for Anomaly Identification}

Conditional Random Fields (CRFs) are \cite{Lafferty2001} discriminative probabilistic graphical models to label data sequences. A CRF simply employs feature functions ($\mathrm{f_j}$) to model the underlying probability distribution of the model. Each feature function can take value of one or zero, since it is an indicator function. Each feature function also has a weight denoted with $\lambda$. Therefore, a CRF is represented by using these feature functions and their corresponding weights, and in order to label a sequence, the normalized product of these feature functions is taken into consideration. Considering other generative models, it has the opportunity to have more relaxed assumptions between the hidden states and the observations, and can have arbitrary number of feature functions in the model. 

\begin{figure}[h]
\begin{center}
     {\includegraphics[scale=0.53]{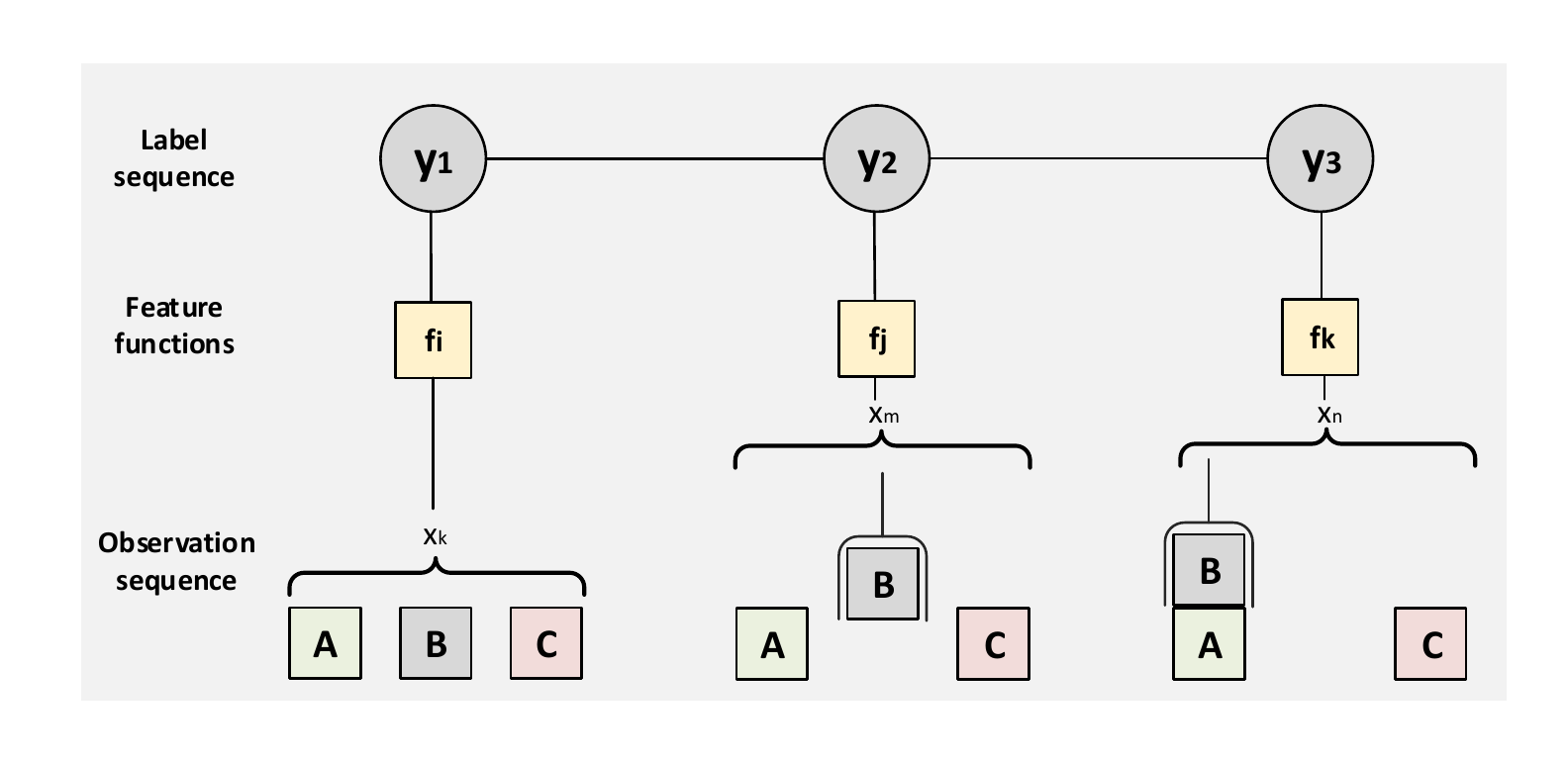}}
\caption{\label{fig:feature-function}An example sequence with feature functions for the block stacking scenario.}
\end{center}
\end{figure}

In Figure \ref{fig:feature-function}, an example sequence for the block stacking scenario with corresponding feature functions is illustrated. In the figure, $y_i$ represents a hidden state that is not known initially, and the aim is to label each such variable with a class. Each observation that the sequence gathers for each state is denoted with $x_k$, $x_m$ and $x_n$. An example feature function is also defined, and it is denoted with $\mathrm{f}_j(y_{i-1},y_{i},x_m,S,i)$.

In this formulation, $j$ stands for the index of the feature function, $y_{i}$ denotes the state that is not directly observable, $S$ is for the whole sequence and $i$ is the index of the state in the sequence. Therefore, this feature function represents the case where the current state $y_{i} = anomaly_k$, the previous state's label is $y_{i-1} = anomaly_n$ and the observation is $obs_m$. Furthermore, $obs$ is an element of the observation set. Note that in this illustration, each feature function is associated with a single observation and state for simplicity. There may be other feature functions that are applicable in the current state as well.

Feature functions are binary functions, their domains include only two values: 0 and 1. Therefore, the feature function illustrated in Figure \ref{fig:feature-function} can be defined as in Equation \ref{f-function}.

\begin{equation}
 \mathrm{f_j}(y_{i-1},y_i,x_i,S,i) = \\
\begin{cases}
1,  & \textit{if } y_{i-1} = anomaly_n \quad \wedge \\ 
&  y_i = anomaly_k \quad \wedge \quad x_i = obs \\
0, & \text{otherwise}
\end{cases}
\label{f-function}
\end{equation}

These feature functions are used with their weights in the model in order to maintain a probability distribution. Therefore, each feature function has its own weight, denoted with $\lambda_j$. Given a sequence, probability of a state is defined as in equation \ref{f-fun}.

\begin{equation}
P(y|x) = \frac{exp[\sum\limits_{j=1}^m \lambda_j \mathrm{f_j}(y_{i-1},y_i,x_i,S,i)]}{\sum \limits_{y^\prime} exp[\sum\limits_{j=1}^m \lambda_j \mathrm{f_j}(y^\prime_{i-1},y^\prime_i,x_i,S,i)]}
\label{f-fun}
\end{equation}

In equation \ref{f-fun}, the nominator part takes into account the score for the specific state considering the feature functions and their weights. On the other hand, the denominator is used for normalization (where $y^\prime$ stands for each hidden state) and is generally called as $Z$.

\section{Experiments}
This section explains the experiment setup, the details on the data collection, training and testing processes, and presents the results.
\subsection{Experimental Setup}


Experiments take place in a laboratory environment that contains objects placed on a table. The following objects are used in the experiments: a powder box, a milk bottle, a plastic grape toy, a pasta box and three cubical blocks. The Baxter robot is used to manipulate these objects in the environment. The robot is tasked to manipulate the objects by the available manipulation actions. Anomaly cases that occur during these scenarios are investigated.

\subsubsection{Data Collection and Annotation}

In the experiments, 120 scenarios where Baxter is tasked to manipulate objects are investigated. 49 of them 
are the scenarios where the target object disappears from the  environment by the manipulation of a human. There are 39 scenarios 
in which the robot is tasked to construct a tower from the objects in the environment and the tower collapses due to unbalanced stacking of objects. Note that in the tower construction scenarios, there exist safe scenarios where the objects are stacked with an offset, however the tower does not collapse as well. The rest of the data set 
includes scenarios where the object is displaced beyond the knowledge of the robot. All of the scenarios are recorded with the sensory data, states and observations. The sampling frequency is at 10 Hz.

Before training the models, the data set is annotated to indicate anomaly occurring times. In order to do so, all records are labeled in such a way that it includes the time stamps on which the anomaly case occurs. Then, the data set is randomly split into training and test sets. For each run, \%80 of the data set is used as training set where the remaining partition is used for testing the performance of the methods. 

\subsubsection{A Temporal Analysis of an Anomaly Case}

\begin{figure*}[h]
    \centering
    \captionsetup[subfigure]{justification=centering}
    \begin{subfigure}[t]{0.3\textwidth}
        \centering
        \includegraphics[width=1.4in]{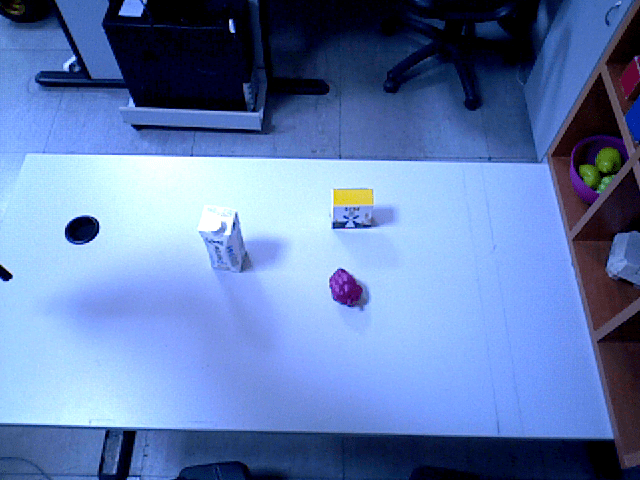}
       \caption{}
    \end{subfigure}%
    ~ 
    \begin{subfigure}[t]{0.3\textwidth}
        \centering
        \includegraphics[width=1.4in]{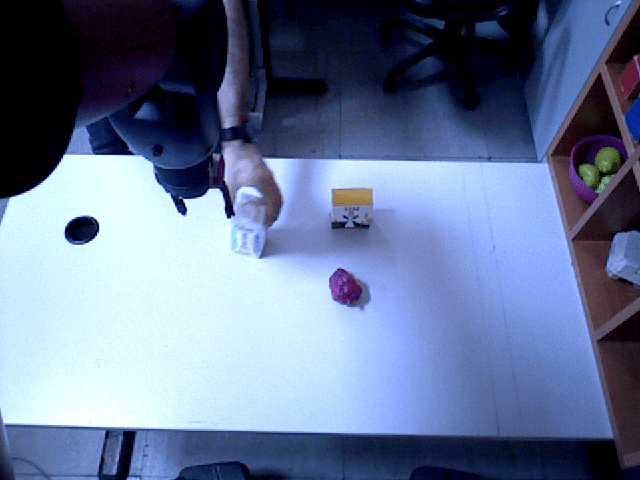}
        \caption{}
    \end{subfigure}
     ~ 
    \begin{subfigure}[t]{0.3\textwidth}
        \centering
        \includegraphics[width=1.4in]{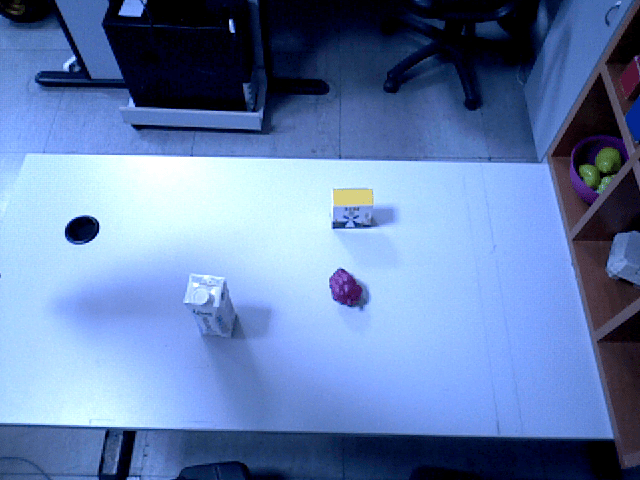}
        \caption{}
    \end{subfigure}
    \caption{\label{external}  A sample anomaly case scenario where the human changes the location of the object which is about to be manipulated. The images are gathered from the RGB-D camera mounted on the Baxter. (a) The robot is tasked to manipulate the milk bottle that exists on the table. (b) While the robot moves through the object to pick it up, the location of the object is changed by a human. (c) This situation leads the robot to an anomaly case, and the robot detects the anomaly after it moves its arm from its point of view. Later on, the identification procedure is activated to find the related cause of the anomaly by investigating the whole execution sequence.}
\end{figure*}

A sample anomaly case is selected to illustrate how sensory data change temporally during execution before the anomaly is detected. Figure \ref{external} illustrates the snapshots from this execution sequence from the view of the robot's RGB-D camera. First, the robot observes the objects in the environment, then it is tasked to manipulate the milk bottle which is initially localized by the vision algorithm. The robot constructs a path plan to reach the object with its arm and moves the arm towards the milk bottle. However, a human intervenes in the scene and changes the location of the bottle while the robot tries to pick it up. The robot detects the anomaly, and moves its arm to its initial home position. After it observes the scene, the milk bottle is recognized in its new location.

\begin{figure*}[h!]
\begin{center}
     {\includegraphics[scale=0.6]{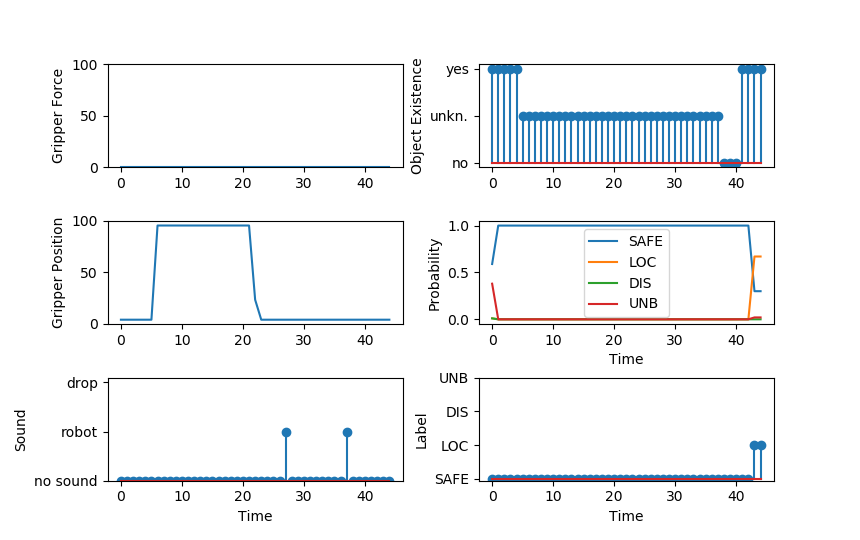}}
\caption{\label{ext-graph}Interpretation of the sensor values gathered by the robot for the change in the object's location and illustration of the robot's belief on each anomaly type.}
\end{center}
\end{figure*}


Figure \ref{ext-graph} shows the interpretation of the sensor values gathered by the robot during such an anomaly case (Figure \ref{external}). The figure includes six plots, and each of them illustrates a different sensor data or interpretation. x axes of all graphs indicate the time in seconds where y axes denote the corresponding value for that feature. The first graph indicates the force in the gripper of the robot. Whenever the gripper is empty (an object is not being hold), it is zero. Otherwise, it produces a value other than zero indicating that there is an object in the gripper. The second graph illustrates the position of the gripper pedals. Whenever they are closed, it is almost zero. Otherwise, depending on the object's size at hand, it has a value other than zero and smaller than 100. The next graph shows the classified sound information, and it is followed by the graph that shows the belief on the existence of an object, depending on the outputs of the vision algorithms. The graph shows the interpretation only for a single object for clarity. The value \textit{yes} indicates that the object is recognized by the vision module, and it is observed by the robot where the value \textit{no} indicates the non-existence of the object. \textit{unkn.} denotes that the visual updates of the robot for that object is suspended due to the object is occluded with the robot's arm. 
The next plot shows the robot's belief on an anomaly case for each class as normalized probabilities. Note that probabilities of the reasons for anomalies are updated at each time step based on the observations. The last graph indicates the label of the state. This is the actual label (ground truth) for the observation which is assigned by annotating the data. \textit{SAFE} denotes the safe state where the state does not include a clue of an anomaly, \textit{LOC} corresponds to the anomaly type where the location of an object is changed, \textit{DIS} is the abbreviation for the cases in which an object which was initially in the scene disappears and \textit{UNB} corresponds to cases where an unstable structure, particularly a tower, is collapsed due to an unbalanced sub-tower that is constructed during a previous action. Initially, the robot observes the objects in the environment. Since the first action is to grasp the object, it opens its gripper and this causes a peak in the gripper position. Note that while the robot's hand occludes the camera, visual updates related to the objects are suspended during that period. Therefore, the belief on the existence of an object is set to \textit{unknown} due to Open World Assumption (OWA). Since the object is transferred to another location on the table, the robot fails to grasp it. After the anomaly is detected, the robot moves its arm to its home position. The location updates on the object model is achieved between 38-40 seconds. 

\subsection{Experimental Results}
In this section, the comparative analysis of the anomaly identification methods is presented. The data set collected for the experiments is evaluated with mainly four methods: LSTM-based, two CRF-based and HMM-based methods. The performance of the CRF-based methods are evaluated with different parameter learning algorithms. The following methods are tested during the analysis.

\begin{itemize}
\item HMM-based identification procedure

\item CRF-based identification procedure with Adaptive Regularization of Weight Vector (AROW) \cite{Crammer2009}

\item CRF-based identification procedure with Limited-memory Broyden-Fletcher-Goldfarb-Shanno (L-BFGS) \cite{Saputro2017}

\item LSTM-based identification procedure

\end{itemize}

The LSTM-based method is applied by using the following parameters: Adam Algorithm \cite{Kingma2014} is used as an optimizer and Cross Entropy Loss is used to calculate the loss during the training phase of the models. Models are trained for 500 epochs with a learning rate of 0.001. 
\begin{figure}[h]
\begin{center}
     {\includegraphics[scale=0.45]{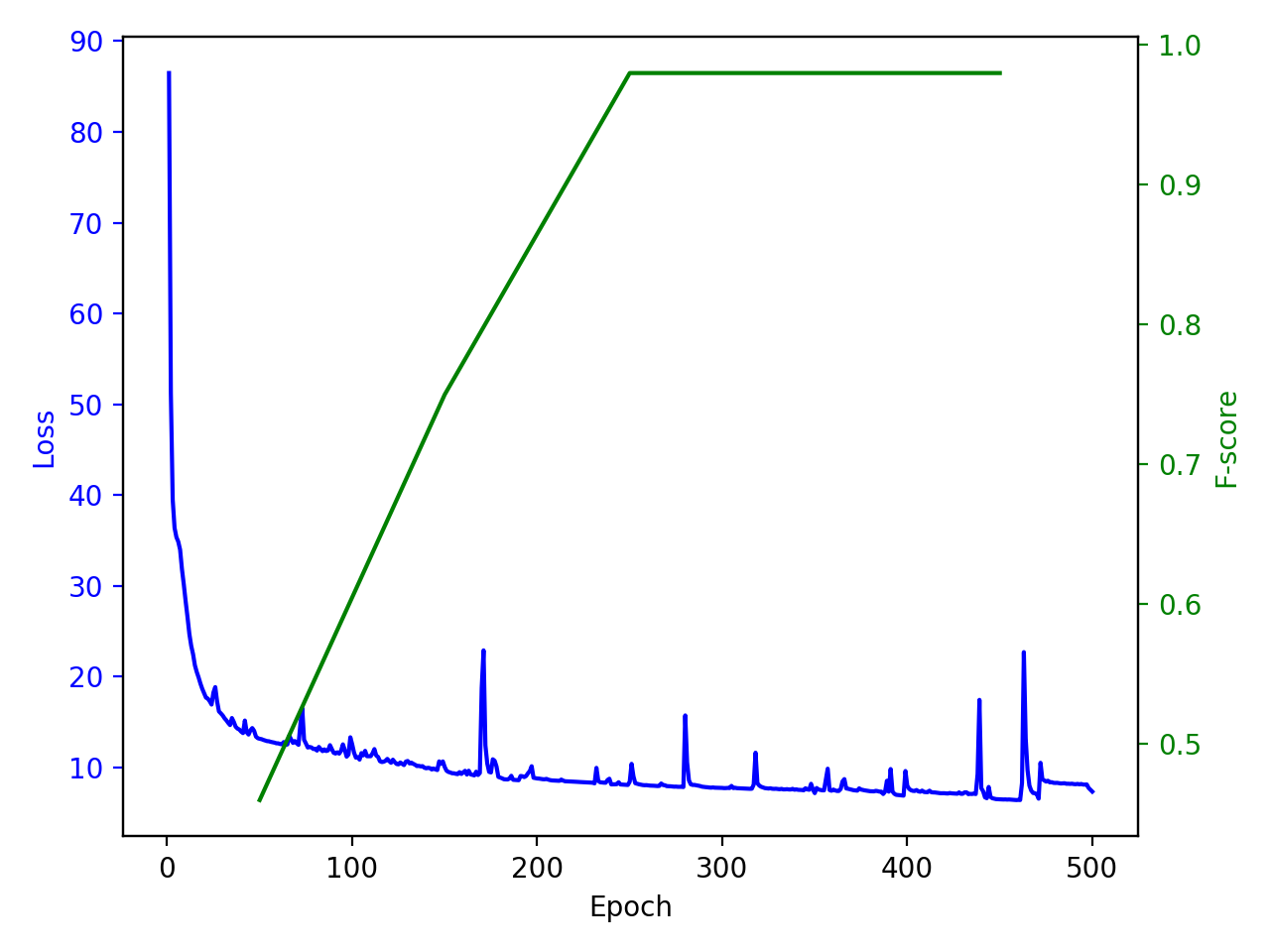}}
\caption{\label{loss}The change on loss (blue) and f-score (green) during the training phase with a learning ratio of 0.001 for LSTM.}
\end{center}
\end{figure}

Figure \ref{loss} illustrates the change on the loss function during a training phase for the LSTM model. As can be seen from the plot, the loss decreases and converges as number of epochs increase. The f-score values for the related models at corresponding epochs are also shown in the plot. It can be seen that after epoch 300, the f-score does not change, and it can be inferred that the trained model converges. 

During the experiments, the following aspects are investigated. First, the performance analysis of the methods is presented in terms of the classification rate of the states, and also confusion matrices are presented. Second, the general classification accuracies of the methods in terms of the overall anomaly identification decision for the scenarios are presented as confusion matrices.

\subsubsection{State Classification Accuracy on the Observation Sequence}

The methods are evaluated in terms of the state classification accuracy at each time step. In this analysis, each observation of the robot is labeled by the methods, and precision, recall and f-score values of the methods are presented.

\begin{table*}[h]
\centering
\resizebox{12cm}{!}{%
\begin{tabular}
{>{}c |
>{}c |
>{}c |
>{}c |
>{}c |
>{}c |
>{}c |
>{}c |
>{}c |
>{}c |
>{}c |
>{}c |}
\cline{3-12}
\multicolumn{2}{c}{\textit{\textbf{}}} & \multicolumn{2}{|c|}{\textbf{Safe}} & \multicolumn{2}{c|}{\textbf{Location}} & \multicolumn{2}{c|}{\textbf{Disappearance}} & \multicolumn{2}{c|}{\textbf{Unbalance}} & \multicolumn{2}{c|}{\textbf{Overall}} \\ \cline{3-12} 
\multicolumn{1}{l}{ } & \multicolumn{1}{l|}{} & \textbf{AVG} & \textbf{STD} & \textbf{AVG} & \textbf{STD} & \textbf{AVG} & \textbf{STD} & \textbf{AVG} & \textbf{STD} & \textbf{AVG} & \textbf{STD} \\ \hline \hline
\multicolumn{1}{|c|}{} & \textbf{HMM} & 0.88 & 0.04 & 0.39 & 0.37 & 0.83 & 0.30 & 0.53 & 0.08 & 0.66 & 0.20 \\ 
\multicolumn{1}{|c|}{} & \textbf{CRF w/ AROW} & 0.92 & 0.05 & 0.77 & 0.31 & 0.74 & 0.22 & 0.80 & 0.15 & 0.81 & 0.18 \\ 
\multicolumn{1}{|c|}{} & \textbf{CRF w/ L-BFGS} & 0.91 & 0.04 & \textbf{1.00} & 0.00 & 0.84 & 0.12 & \textbf{0.99} & 0.01 & 0.94 & 0.04 \\ 
\multicolumn{1}{|c|}{\multirow{-4}{*}{\textbf{Precision}}} & \textbf{LSTM} & \textbf{0.96} & 0.02 & \textbf{1.00} & 0.00 & \textbf{0.95} & 0.07 & 0.94 & 0.05 & \textbf{0.96} & 0.05 \\ \hline
\hline
\multicolumn{1}{|c|}{} & \textbf{HMM} & 0.85 & 0.16 & 0.65 & 0.41 & 0.59 & 0.36 & 0.64 & 0.32 & 0.64 & 0.25 \\ 
\multicolumn{1}{|c|}{} & \textbf{CRF w/ AROW} & 0.97 & 0.04 & 0.75 & 0.33 & 0.44 & 0.36 & 0.60 & 0.26 & 0.69 & 0.25 \\ 
\multicolumn{1}{|c|}{} & \textbf{CRF w/ L-BFGS} & \textbf{1.00} & 0.00 & 0.73 & 0.22 & 0.78 & 0.21 & 0.63 & 0.16 & 0.78 & 0.15 \\ 
\multicolumn{1}{|c|}{\multirow{-4}{*}{\textbf{Recall}}} & \textbf{LSTM} & 0.99 & 0.01 & \textbf{0.95} & 0.02 & \textbf{0.95} & 0.05 & \textbf{0.81} & 0.08 & \textbf{0.93} & 0.04 \\ \hline
\hline
\multicolumn{1}{|c|}{} & \textbf{HMM} & 0.86 & 0.12 & 0.41 & 0.31 & 0.64 & 0.32 & 0.50 & 0.07 & 0.60 & 0.21 \\ 
\multicolumn{1}{|c|}{} & \textbf{CRF w/ AROW} & 0.94 & 0.02 & 0.68 & 0.32 & 0.46 & 0.25 & 0.64 & 0.21 & 0.68 & 0.20 \\ 
\multicolumn{1}{|c|}{} & \textbf{CRF w/ L-BFGS} & 0.95 & 0.02 & 0.83 & 0.17 & 0.79 & 0.12 & 0.76 & 0.12 & 0.83 & 0.11 \\ 
\multicolumn{1}{|c|}{\multirow{-4}{*}{\textbf{F-Score}}} & \textbf{LSTM} & \textbf{0.97} & 0.01 & \textbf{0.98} & 0.01 & \textbf{0.95} & 0.04 & \textbf{0.87} & 0.05 & \textbf{0.94} & 0.03 \\ \hline
\end{tabular}
}
\caption{\label{overall-perf}Overall performance comparison and analysis of the anomaly identification methods.}
\end{table*}

Table \ref{overall-perf} presents the statistical analysis of the methods. Each method is run 10 times on randomly partitioned data sets (80\% training and 20\% testing), and the average and standard deviation values are presented for precision, recall and f-score metrics. Each column in the table presents the results for a distinct class: success and three distinct anomaly classes. The table is divided into three parts. Each part presents the analysis of a different performance measure, precision, recall and f-score, consecutively. The results of each method are given as a separate line within each part.

As can be seen from the table, except the cases that include unbalance anomalies, LSTM-based method overcomes HMM-based and CRF-based methods in terms of precision. Between the CRF-based methods, the algorithm trained with L-BFGS outperforms the others in terms of the overall average precision for the anomaly classes. HMM-based method is not able to distinguish between location change and disappearance anomalies efficiently. Similarly for recall and f-score measures, LSTM-based method again outperforms the other methods for the anomaly classes. Moreover, it has also lower standard deviation values compared to the other methods. The main reason behind this is the ability of dealing with propagating the effects of an observation that is gathered before the upcoming steps (vanishing gradients problem). For example, consider the unbalance anomaly. Anomaly indicators are not perceived at a time simultaneously, but gathered sequentially at distinct time steps. That is, first the robot receives a collapse sound, then it perceives the visual information of a cluttered environment. Furthermore, some objects may not be recognized after the collapse. Even though objects may be recognized, the locations of them are changed due to the collapse. The indicators of all anomaly types are all included in that type of scene: location change, object disappearance and unbalance. However, LSTM-based model is able to relate the observation sequence efficiently in a temporal manner compared to the other methods, and is able to distinguish among anomaly types successfully. Consider another scenario where the object is taken out of the environment (disappearance anomaly). The robot can relate these sequential observations during that scenario: the object was observable before, an action on that object is not applied and now it is disappeared. The LSTM-based approach is capable of relating these kind of sequences better compared to the HMMs and CRFs. 

Another interpretation that can be made from the results is the relative low accuracy for the unstable anomalies compared to the other anomaly cases for all methods. After a tower collapses, most of the time, the robot perceives a cluttered environment. Since this is a complex scene, sensor accuracies especially for visual modality decreases. Therefore, the f-score of this anomaly is relatively low compared to the other anomaly classes.

\begin{figure*}[h]
    \begin{subfigure}[t]{0.5\textwidth}
        \hfill
         {\includegraphics[scale=0.38]{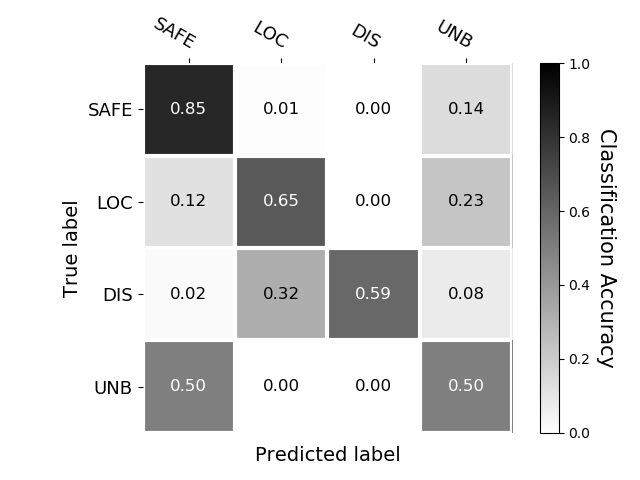}}
\caption{\label{heat-map-confusion-hmm}HMM-based method.}
    \end{subfigure}%
    \begin{subfigure}[t]{0.5\textwidth}
        \hfill
         {\includegraphics[scale=0.38]{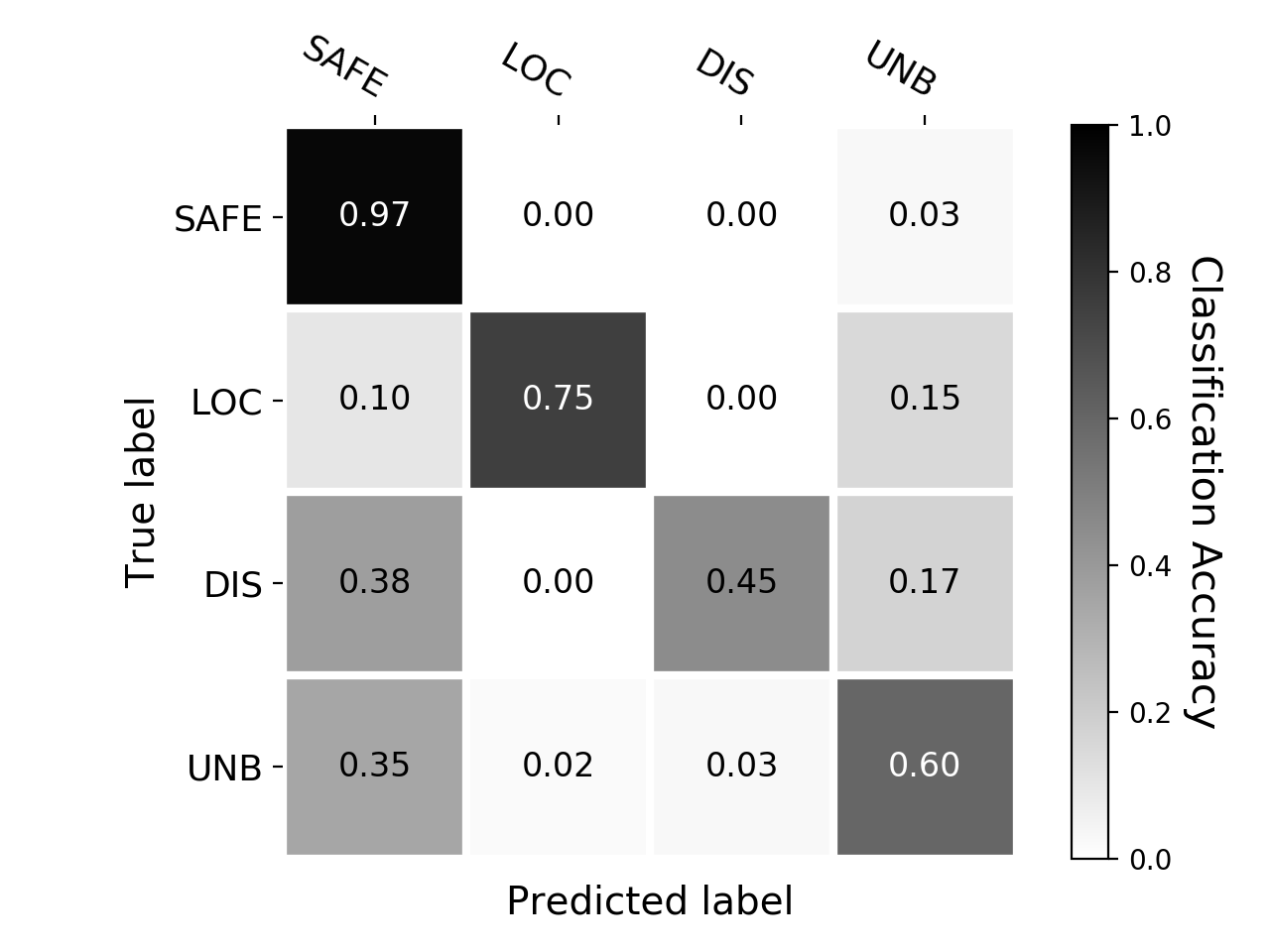}}
\caption{\label{heat-map-confusion-arow}CRF-based method (with AROW).}
    \end{subfigure}%
     \hfill
    \begin{subfigure}[t]{0.5\textwidth}
        \hfill
        {\includegraphics[scale=0.38]{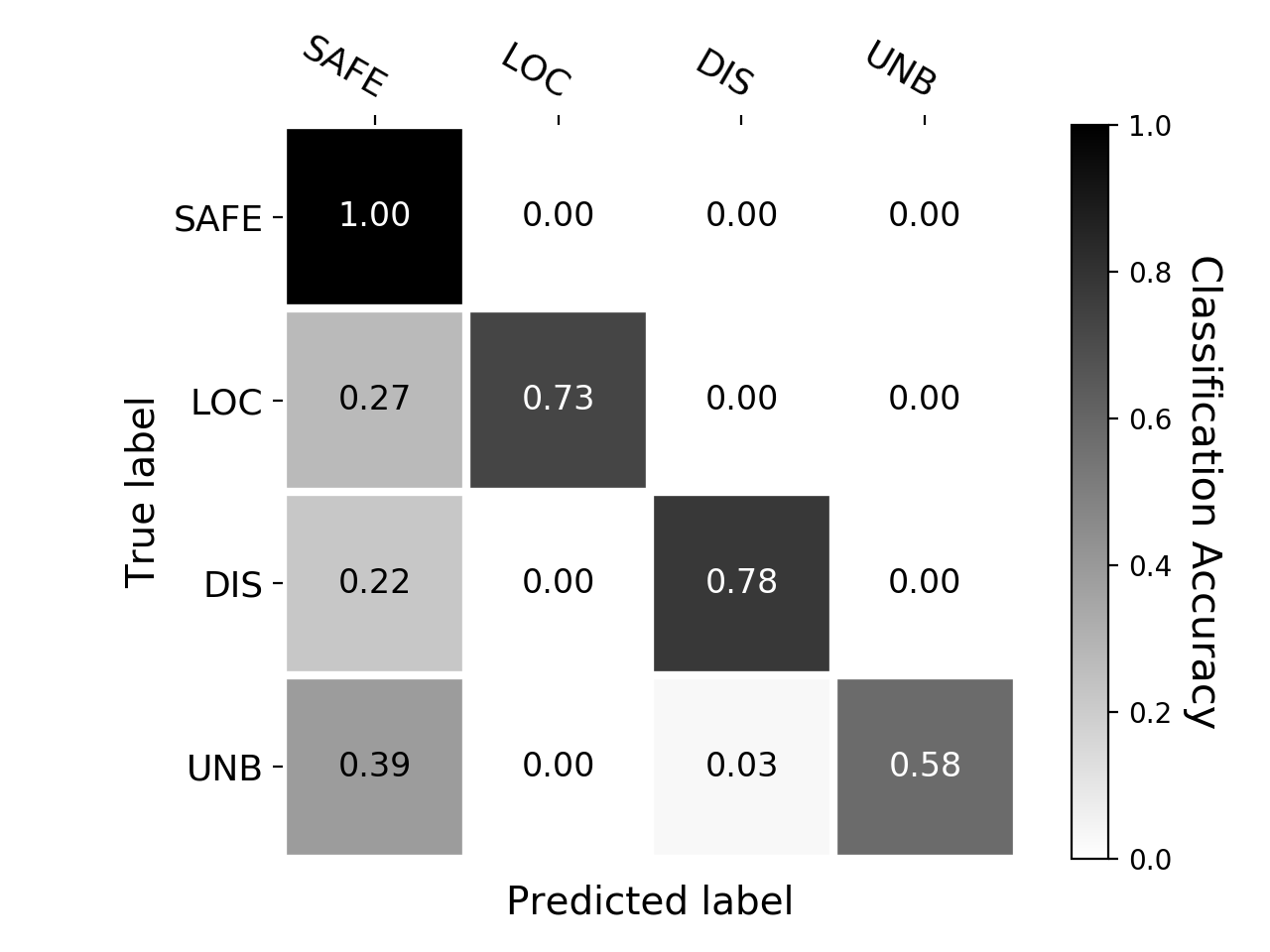}}
\caption{\label{heat-map-confusion-lbfgs}CRF-based method (with L-BFGS).}
    \end{subfigure}
    \begin{subfigure}[t]{0.5\textwidth}
        \hfill
        {\includegraphics[scale=0.38]{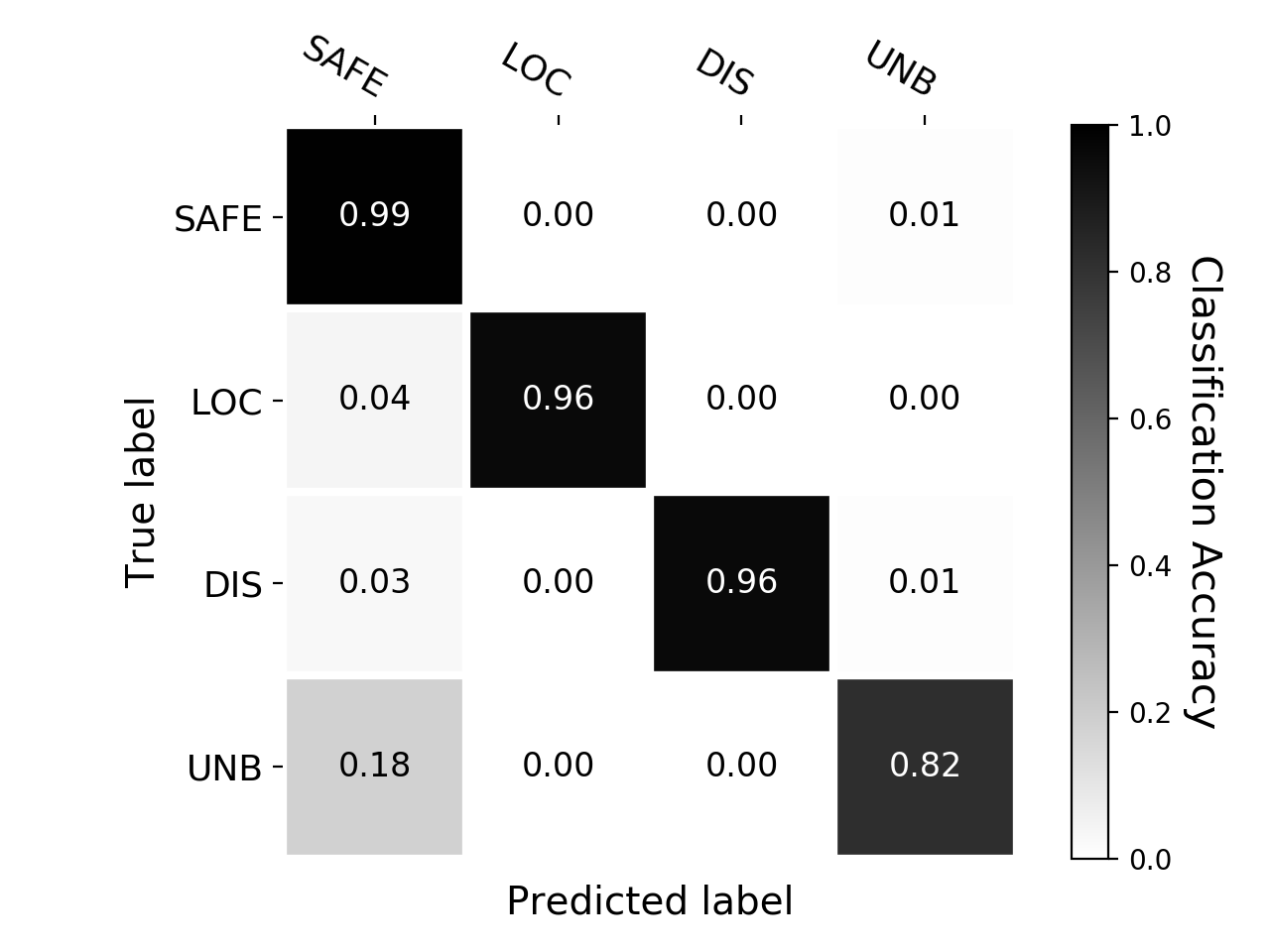}}
\caption{\label{heat-map-confusion-lstm} LSTM-based method.}
    \end{subfigure}%
   
    \caption{\label{case-classification-confusion-heat}  Confusion matrices for the methods for the state classification task.}
\end{figure*}

Figure \ref{case-classification-confusion-heat} presents confusion matrices for the same task presented in this section. Figure \ref{heat-map-confusion-hmm} shows the confusion matrix for the HMM-based method where Figure \ref{heat-map-confusion-arow} shows the confusion matrix for the CRF-based method with AROW learning. The other sub-figures show the results for CRF-based method with L-BFGS and LSTM-based methods, respectively. It can be seen from the matrix that the CRF-based and HMM-based  methods confuse anomaly classes most of the time. For example, for the anomaly scenarios where an object is taken out of the scene (disappearance anomalies), CRF-based method with AROW learning confuses this anomaly with success cases and unbalance anomalies. Similarly, the other CRF-based method confuses unbalance anomalies with success cases and disappearance anomalies, and confuses location change anomalies with success cases (Figure \ref{heat-map-confusion-lbfgs}). However, LSTM-based anomaly identification method can successfully classify anomaly states with a minimum of 82\% rate with minimizing the confusion of the anomaly classes (Figure \ref{heat-map-confusion-lstm}). 

\subsubsection{Overall Anomaly Identification Accuracy}
In this section, case (scenario) classification accuracy is presented. After an anomaly is detected, the full observation sequence is tagged with one of the identification methods. That is, each observation in the sequence is classified with one of the anomaly classes. However, a single final overall decision should be given for that scenario to decide on the anomaly type that is occurred.  To do so, majority voting is applied to end up with a final decision for an anomaly case. After labelling each observation of the case, the maximum number of observations labeled with a specific anomaly is accepted as the identification of that case.

\begin{figure*}[h]
    \begin{subfigure}[t]{0.5\textwidth}
         \hfill
         {\includegraphics[scale=0.35]{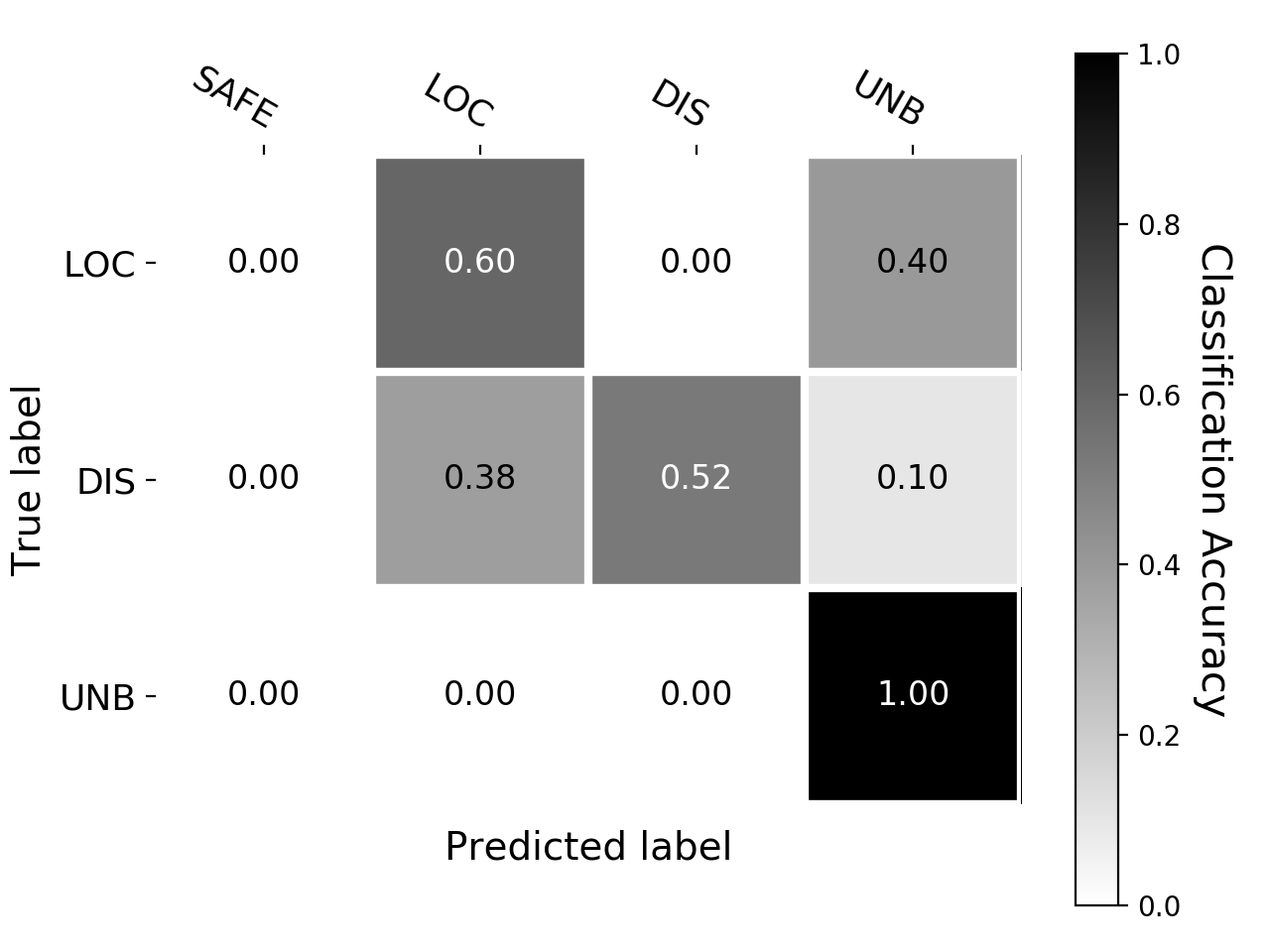}}
\caption{\label{heat-map-hmm}HMM-based method.}
    \end{subfigure}%
    \begin{subfigure}[t]{0.5\textwidth}
         \hfill
         {\includegraphics[scale=0.35]{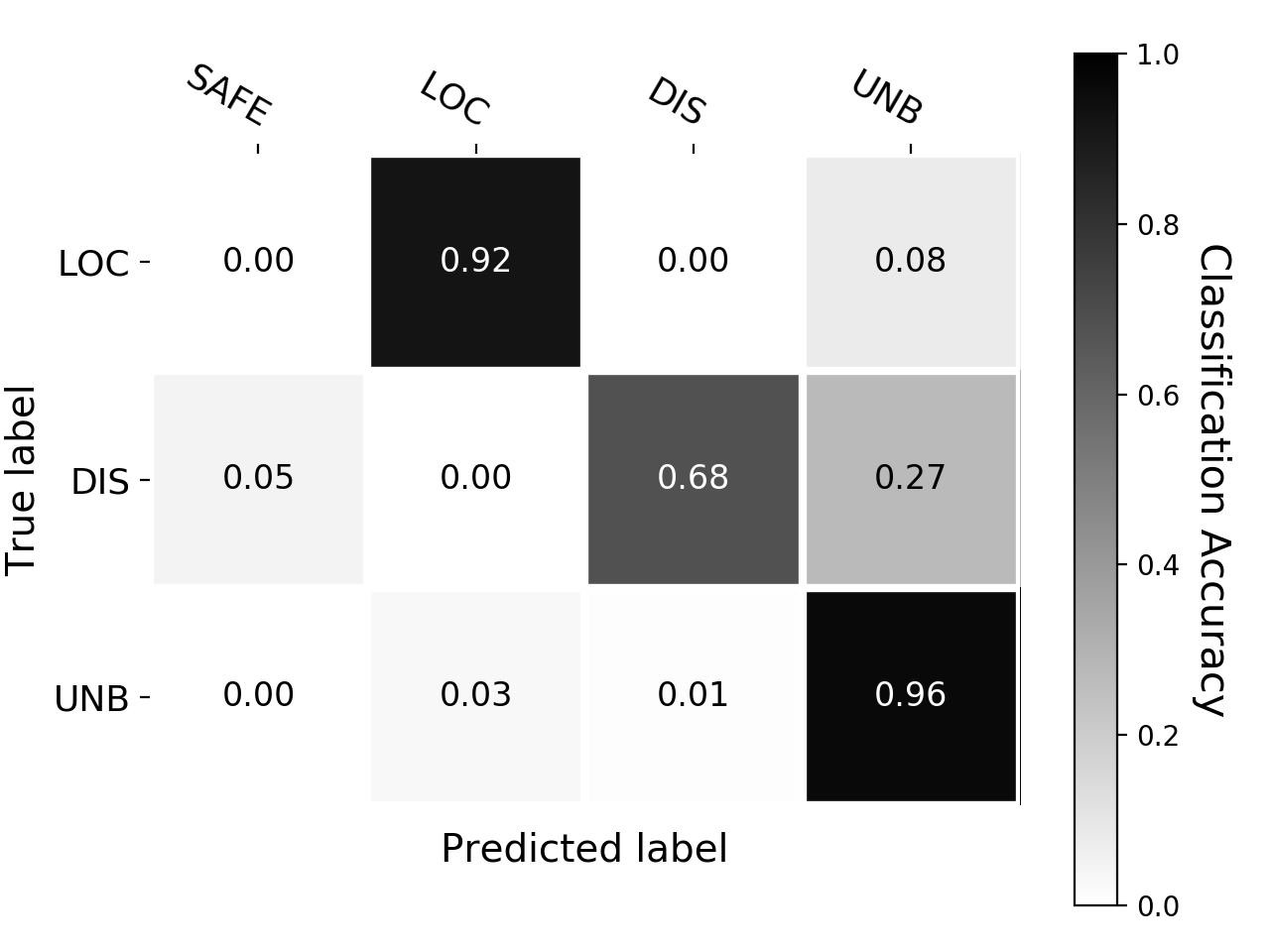}}
\caption{\label{heat-map-arow}CRF-based method (with AROW).}
    \end{subfigure}%
    \hfill
    \begin{subfigure}[t]{0.5\textwidth}
         \hfill
          {\includegraphics[scale=0.35]{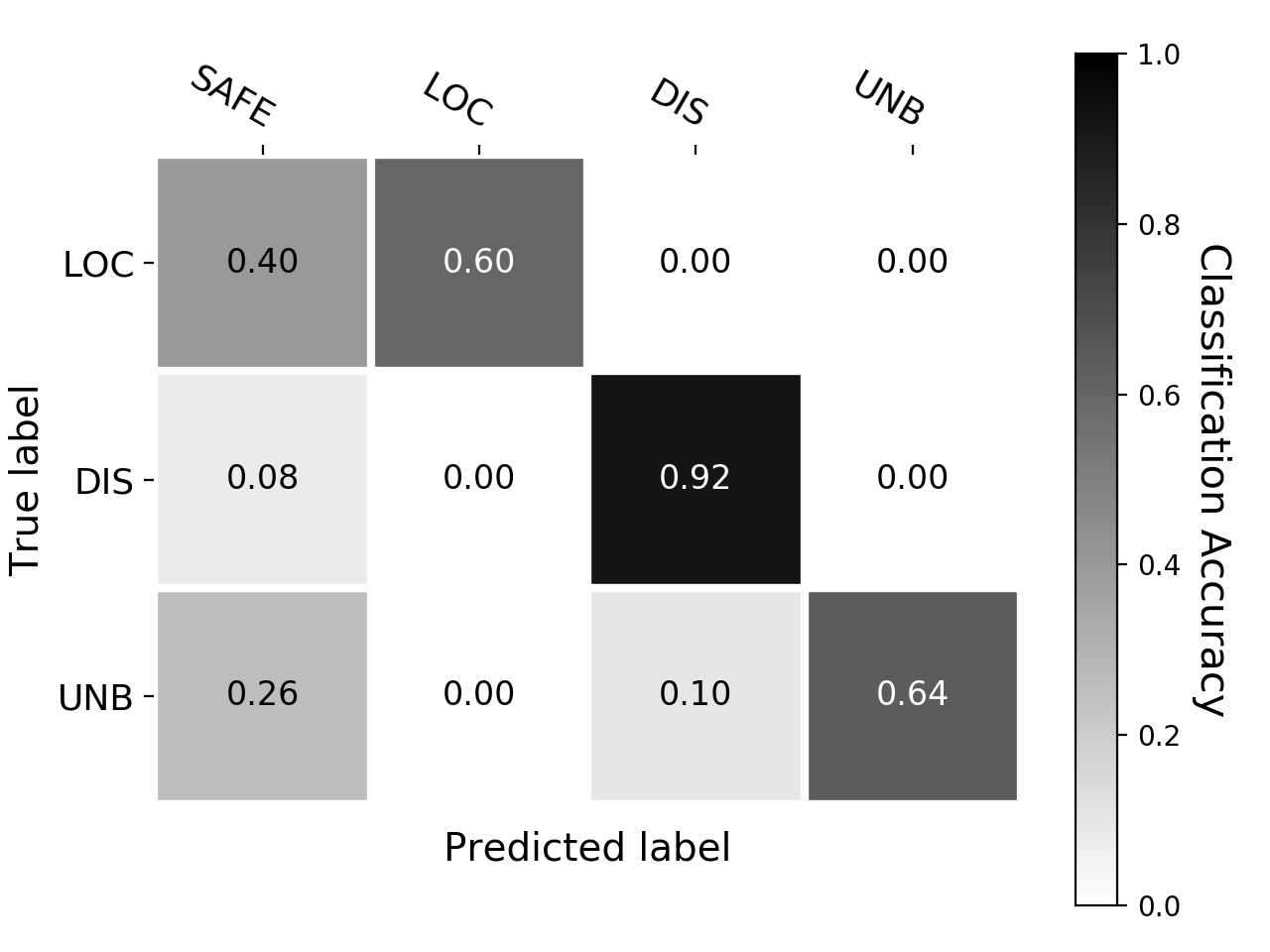}}
\caption{\label{heat-map-lbfgs}CRF-based method (with L-BFGS).}
    \end{subfigure}
    \begin{subfigure}[t]{0.5\textwidth}
         \hfill
        {\includegraphics[scale=0.35]{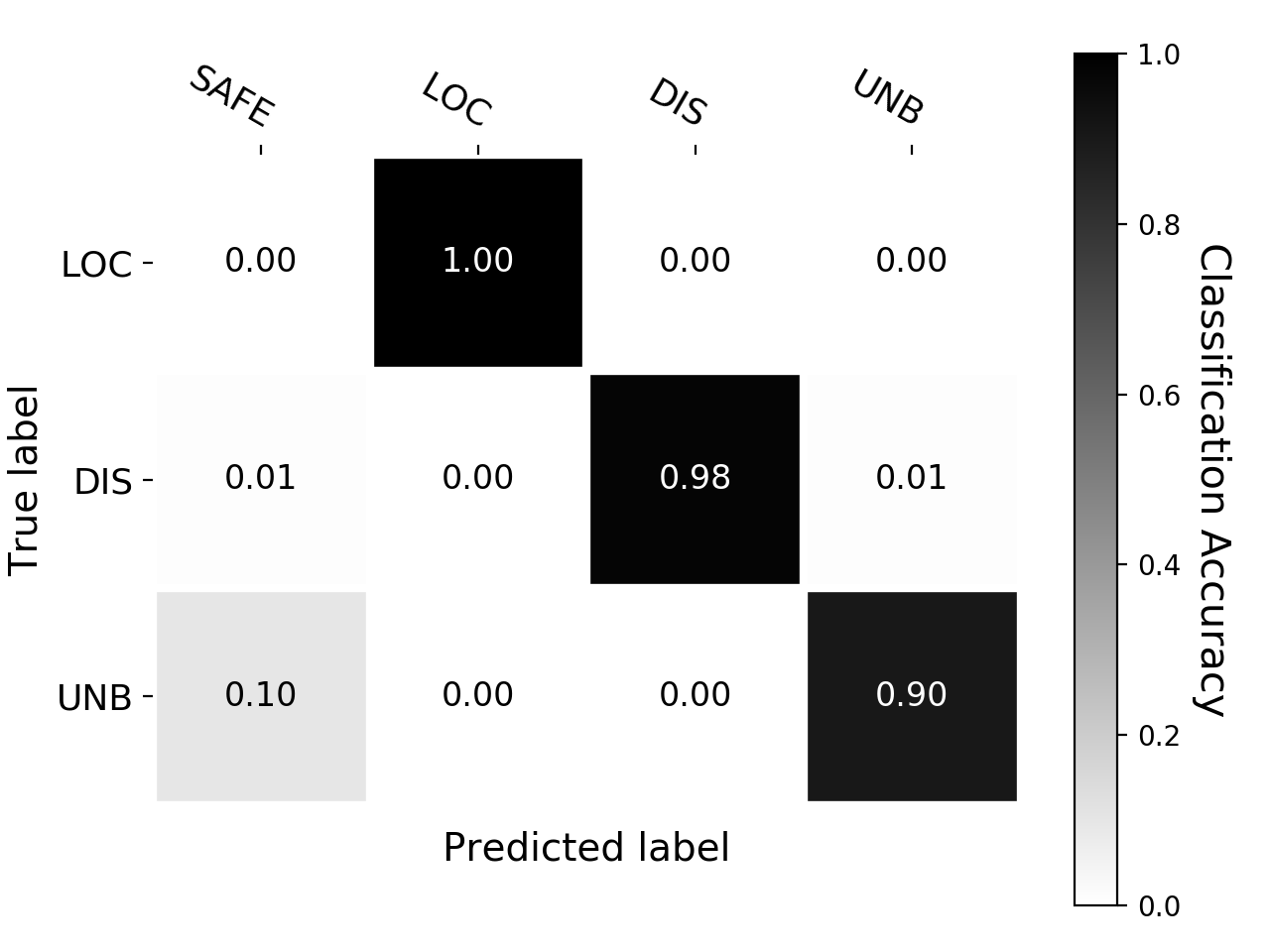}}
\caption{\label{heat-map-lstm} LSTM-based method.}
    \end{subfigure}%
   
    \caption{\label{case-classification-heat}  Case classification accuracy for the methods in terms of the number of scenarios.}
\end{figure*}

Figure \ref{case-classification-heat} illustrates the confusion matrices of the methods in the case classification task. The first sub-figure (Figure \ref{heat-map-hmm}) presents the confusion matrix for the HMM-based method. It confuses the location change and disappearance anomalies most of the time since their indicators are quite similar (both related to location). On the other hand, it can distinguish collapse anomalies better.
The second sub-figure (Figure \ref{heat-map-arow}) presents the confusion matrix for the CRF-based method with AROW learning algorithm. Although it has lower performance in Table \ref{overall-perf}, for case classification task, it performs better, especially for location and unbalance scenarios. The main reason behind this can be explained as follows. The low state classification performance of the CRF-based method with AROW learning means that it confuses the labels most of the time during the classification process. That is, the classification can not reach a consensus on a label, it may switch among labels during the labeling process of an anomaly case. However, most of the time the highest number of labels comes from the true anomaly label. Since majority voting is used to make an overall decision, this leads to a performance increase for the location and unbalance anomalies. However, for the other anomaly type, this method confuses more and the performance is low. As can be seen from the confusion matrix in Figure \ref{heat-map-lbfgs}, the instability of the labeling during classification is valid for the CRF-based method with L-BFGS learning, as well. Figure \ref{heat-map-lstm} illustrates the confusion matrix for the LSTM-based method. For all the anomaly cases, it performs 0.90 or above and it does not confuse much while classifying the observations. Note that the label \textit{SAFE} does not exist in Figure \ref{case-classification-heat} (unlike Figure \ref{case-classification-confusion-heat}) since only the cases where an anomaly occurs are used in the test set, and in this task, a single general decision about the case to explain the anomaly is given, instead of classifying each observation in the corresponding scenario.

\section{Discussion}
 
 This section presents discussions in terms of important aspects of the handled problem and the methodologies presented in this paper. Each aspect is discussed under a distinct subsection.  

\subsection{Variety of Sensors}
An important aspect of the problem presented in this paper is the variety of the sensors that the robot is equipped with. Since the anomaly identification algorithm is mainly observing and interpreting the indicators of anomalies, the performance of the algorithms is correlated with the amount of the data that are gathered from the environment. For example, visual information helps the robot with the existence information of the objects where the microphone provides auditory information about the scene. Furthermore, the robot can be capable of identifying failures related to olfactory anomalies, such as overcooking a meal if it has a smell sensor.

\subsection{Accuracy of Sensors}
For the anomaly identification problem, the most crucial challenge that needs to be handled is the processing and fusing the sensory data in an appropriate manner. Therefore, it is clear that the performance of the algorithms is highly dependent on the accuracy of the sensory data. 

\subsection{Unknown Anomalies}
The methods presented in this paper are model-based methods. First, the models are trained, and they are used for identifying the reasons behind the anomaly cases. However, in case of an anomaly that the robot is not familiar with, the methods would classify the case by considering the common indicators of the anomaly, and would select the anomaly type that has similar or common indicators with the actual anomaly. In order to handle such cases effectively, the methods can be extended in such a way that they have the capability of classifying such unknown anomalies. Intervention of a human may be included in the process to assign a label to the anomaly that the robot encounters with for the first time. Then, an online learning procedure of the model can take place to learn unknown anomalies. 

\subsection{Learning from Anomalies}
 There may be cases where an object may cause an anomaly persistently. In such cases, a learning algorithm \cite{Karapinar2015} can associate this anomaly context with the object and come up with the relevant hypothesis for it. Assume that the robot always places an object in an unstable manner on another object while constructing a tower. After identifying the anomaly, the robot may execute a learning procedure that is able to relate anomaly contexts with objects. With such a hypothesis, for example, the robot may refuse using this object for tower construction for upcoming tasks, if the anomaly context indicates so.

\section{Conclusions and Future Work}
In this paper, we proposed and analyzed an LSTM-based method in order to explain and reveal underlying reasons behind anomaly cases that occur during the execution. The proposed method fuses various sensor modalities in order to explain such cases. The method is also compared with HMM-based and CRF-based methods, and a comparative analysis is presented. Various scenarios that include anomaly cases are investigated on a Baxter robot in an unstructured environment. The results indicate that LSTM-based method outperforms the other methods with a classification rate of 94\% due to its capability of dealing with the vanishing gradient problem. We believe that this study presents an important first step to equip robots with awareness procedures for safety in unstructured environments. Future work includes extending the method so that it is also capable of dealing with unknown anomalies that the robot has not encountered before.

\section{Acknowledgements}
Authors would also like to thank Arda Inceoglu for his beneficial suggestions on this research.

\bibliography{identification.bib}

\end{document}